\documentclass{article}

\PassOptionsToPackage{numbers, compress}{natbib}
\usepackage[final]{neurips_2023}

\usepackage[utf8]{inputenc} 
\usepackage[T1]{fontenc}    
\usepackage{hyperref}       
\usepackage{url}            
\usepackage{booktabs}       
\usepackage{amsfonts}       
\usepackage{nicefrac}       
\usepackage{microtype}      
\usepackage{xcolor}         
\usepackage{graphicx}
\usepackage{amsmath}
\usepackage{amssymb}
\usepackage{subcaption}
\usepackage{wrapfig}
\usepackage{pifont}
\newcommand{\cmark}{\ding{51}}
\newcommand{\xmark}{\ding{55}}

\newcommand{\ours}{XAGen}
\newcommand{\ie}{\emph{i.e.}}
\newcommand{\eg}{\emph{e.g.}}

\definecolor{pearThree}{HTML}{E74C3C}
\definecolor{pearDark}{HTML}{2980B9}
\definecolor{pearDarker}{HTML}{1D2DEC}

\hypersetup{
	colorlinks,
	citecolor=pearDark,
	linkcolor=pearThree,
    breaklinks=true,
	urlcolor=pearDarker}

\title{XAGen: 3D Expressive Human Avatars Generation}

\author{%
  Zhongcong Xu\\
  Show Lab\\
  National University of Singapore \\
  \texttt{zhongcongxu@u.nus.edu} \\
  \And
  Jianfeng Zhang \\
  ByteDance \\
  \texttt{jianfengzhang@bytedance.com} \\
  \And
  Jun Hao Liew \\
  ByteDance \\
  \texttt{junhao.liew@bytedance.com} \\
  \And
  Jiashi Feng \\
  ByteDance \\
  \texttt{jshfeng@bytedance.com} \\
  \And
  Mike Zheng Shou \thanks{Corresponding author} \\
  Show Lab\\
  National University of Singapore \\
  \texttt{mike.zheng.shou@gmail.com}
}

\begin{document}

\maketitle

\begin{abstract}
Recent advances in 3D-aware GAN models have enabled the generation of realistic and controllable human body images. However, existing methods focus on the control of major body joints, neglecting the manipulation of expressive attributes, such as facial expressions, jaw poses, hand poses, and so on. In this work, we present \ours{}, the first 3D generative model for human avatars capable of expressive control over body, face, and hands. To enhance the fidelity of small-scale regions like face and hands, we devise a multi-scale and multi-part 3D representation that models fine details. Based on this representation, we propose a multi-part rendering technique that disentangles the synthesis of body, face, and hands to ease model training and enhance geometric quality. Furthermore, we design multi-part discriminators that evaluate the quality of the generated avatars with respect to their appearance and fine-grained control capabilities. Experiments show that \ours{} surpasses state-of-the-art methods in terms of realism, diversity, and expressive control abilities. Code and data will be made available at \href{https://showlab.github.io/xagen}{https://showlab.github.io/xagen}.
\end{abstract}

\section{Introduction}

3D avatars present an opportunity to create experiences that are exceptionally authentic and immersive in telepresence~\cite{chu2020expressive}, augmented reality (AR)~\cite{grauman2022ego4d}, and virtual reality (VR)~\cite{remelli2022drivable}. These applications~\cite{alexander2010digital,romero2017embodied,bagautdinov2021driving,liu2021spatt} require the capture of human expressiveness, including poses, gestures, expressions, and others, to enable photo-realistic generation~\cite{xu2022pv3d,zhang2023magicavatar}, animation~\cite{siarohin2021motion}, and interaction~\cite{liu2023hosnerf} in virtual environments. 

Traditional methods~\cite{collet2015high,xiang2021modeling,beeler2011high,ghosh2011multiview,guo2019relightables} typically create virtual avatars based on template registration or expensive multi-camera light stages in well-controlled environments.
Recent efforts~\cite{zhang2023avatargen,noguchi2022unsupervised,bergman2022generative,hong2023evad,dong2023ag3d} have explored the use of generative models to produce 3D human bodies and clothing based on input parameters, such as SMPL~\cite{loper2015smpl}, without the need of 3D supervision. 
Despite these advancements, current approaches are limited in their ability to handle expressive attributes of the human body, such as facial expressions and hand poses, as they primarily focus on body pose and shape conditions. Yet, there exist scenarios where fine-grained control ability is strongly desired, \eg, performing social interactions with non-verbal body languages in Metaverse, or driving digital characters to talk with various expressions and gestures, \textit{etc.} Due to the lack of comprehensive modeling of the full human body, existing approaches~\cite{noguchi2022unsupervised,bergman2022generative,hong2023evad} fail to provide control ability beyond the sparse joints of major body skeleton, leading to simple and unnatural animation.

\begin{figure}[t]
\centering
\includegraphics[width=0.77\textwidth]{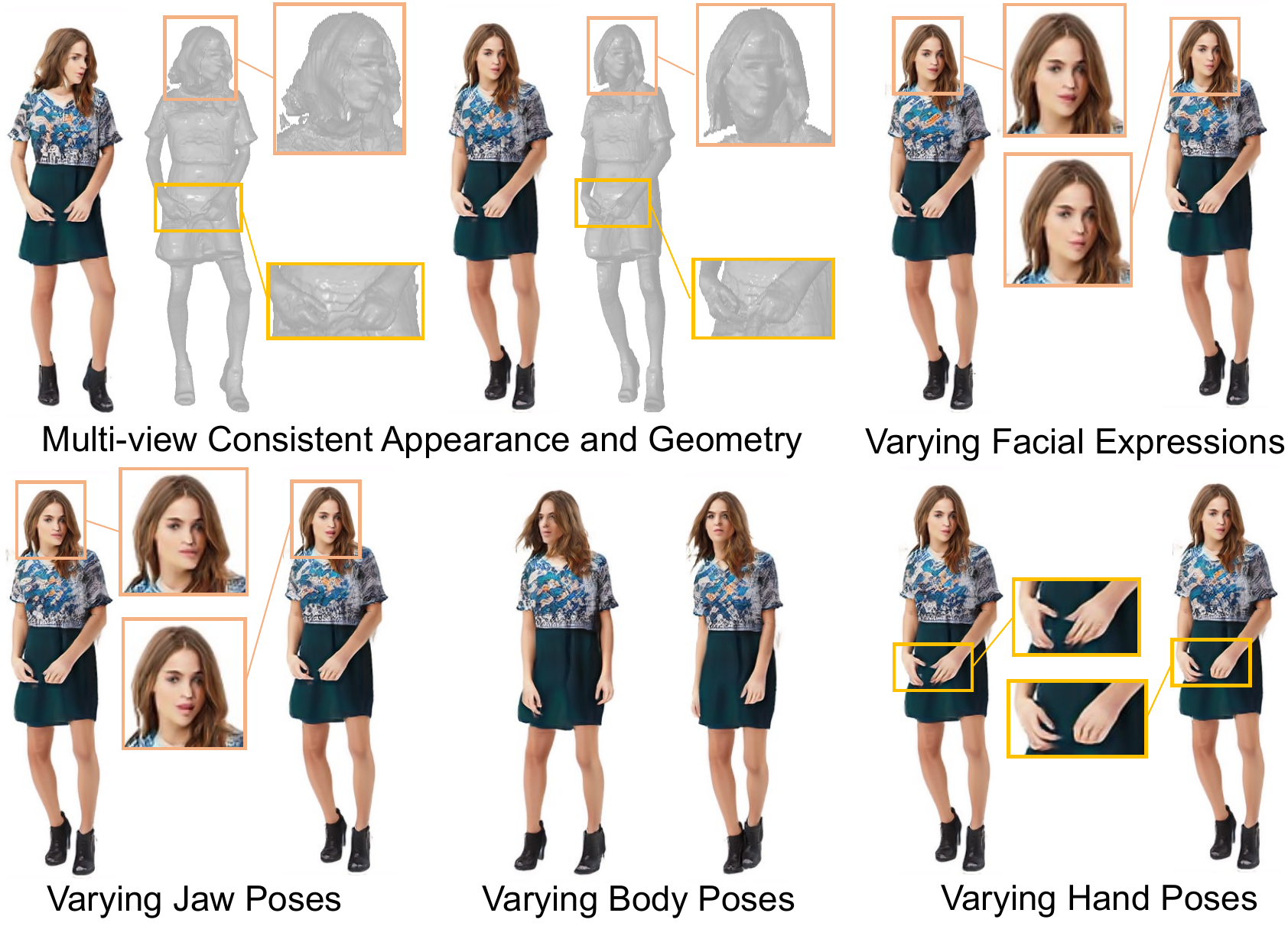}
\caption{\ours{} can synthesize realistic 3D avatars with detailed geometry, while providing disentangled control over expressive attributes, \ie, facial expressions, jaw, body, and hand poses.
}
\label{fig:teaser}
\vspace{-3mm}
\end{figure}

In this work, our objective is to enhance the fine-grained control capabilities of GAN-based human avatar generation model. To achieve this, we introduce the first e{\bf X}pressive 3D human {\bf A}vatar {\bf Gen}eration model ({\bf \ours{}}) that can (1) synthesize high-quality 3D human avatars with diverse realistic appearances and detailed geometries; (2) provide independent control capabilities for fine-grained attributes, including body poses, hand poses, jaw poses, shapes, and facial expressions.

\ours{} is built upon recent unconditional 3D-aware generation models for static images~\cite{chan2022efficient,or2022stylesdf}. One straightforward approach to implement fully animatable avatar generation is extending 3D GAN models to condition on expressive control signals, such as SMPL-X~\cite{smplx}. Though conceptually simple, such a direct modification of conditioning signal cannot guarantee promising appearance quality and control ability, particularly for two crucial yet challenging regions, \ie, the face and hands. This is because (1) Compared with body, face and hands contain similar or even more articulations. In addition, their scales are much smaller than arms, torso, and legs in a human body image, 
which hinders the gradient propagation from supervision. (2) Face and hands are entangled with the articulated human body and thus will be severely affected by large body pose deformation, leading to optimization difficulty when training solely on full-body image collections.

To address the above challenges, we decompose the learning process of body, face, and hands by adopting a multi-scale and multi-part 3D representation and rendering multiple parts independently using their respective observation viewpoints and control parameters. The rendered images are passed to multi-part discriminators, which provide multi-scale supervision during the training process. 
With these careful designs, \ours{} can synthesize photo-realistic 3D human avatars that can be animated effectively by manipulating the corresponding control parameters for expressions and poses, as depicted in Figure~\ref{fig:teaser}. 
We conduct extensive experiments on a variety of benchmarks~\cite{fu2022stylegan,zablotskaia2019dwnet,dong2019towards,liu2016deepfashion}, demonstrating the superiority of \ours{} over state-of-the-arts in terms of appearance, geometry, and controllability. Moreover, \ours{} supports various downstream applications such as text-guided avatar creation and audio-driven animation, expanding its potential for practical scenarios.

Our contributions are three-fold: (1) To the best of our knowledge, \ours{} is the first 3D GAN model for fully animatable human avatar generation. (2) We propose a novel framework that incorporates multi-scale and multi-part 3D representation together with multi-part rendering technique to enhance the quality and control ability, particularly for the face and hands. (3) Experiments demonstrate \ours{} surpasses state-of-the-art methods in terms of both quality and controllability, which enables various downstream applications, including text-guided avatar synthesis and audio-driven animation. 





\section{Related work}
{\bf Generative models for avatar creation.} Generative models~\cite{karras2019style,karras2021alias,rombach2022high} have demonstrated unprecedented capability for synthesizing high-resolution photo-realistic images. Building upon these generative models, follow-up works~\cite{chan2022efficient,or2022stylesdf,shi2021lifting,wang2022rodin,Xu_2023_CVPR_OmniAvatar} have focused on extending 2D image generation to the 3D domain by incorporating neural radiance field~\cite{mildenhall2020nerf} or differentiable rasterization~\cite{kato2018neural}. Although enabling 3D-aware generation, these works fail to provide control ability to manipulate the synthesized portrait images. To address this limitation, recent research efforts~\cite{xu2023omniavatar,hong2023evad,noguchi2022unsupervised,zhang2023avatargen,sun2022next3d,dong2023ag3d,zhang2023getavatar} have explored animatable 3D avatar generation leveraging parametric models for face~\cite{FLAME} and body~\cite{loper2015smpl}. These works employ inverse ~\cite{lewis2000pose} or forward~\cite{chen2021snarf} skinning techniques to control the facial attributes or body poses of the generated canonical avatars~\cite{zhang2023avatargen,zhang2023getavatar}. For human body avatars, additional challenges arise due to their articulation properties. Consequently, generative models for human avatars have explored effective 3D representation designs. Among them, ENARF~\cite{noguchi2022unsupervised} divides an efficient 3D representation~\cite{chan2022efficient} into multiple parts, with each part representing one bone. EVA3D~\cite{hong2023evad} employs a similar multi-part design by developing a compositional neural radiance field. Despite enabling body control, such representation fails to generate the details of human faces or hands since these parts only occupy small regions in the human body images.


Our method differs in two aspects. First, existing works can either control face or body, whereas ours is the first 3D avatar generation model with simultaneous fine-grained control over the face, body, and hands. Second, we devise a multi-scale and multi-part 3D representation, allowing for generating human body with high fidelity even for small regions like face and hands.

{\bf Expressive 3D human modeling.} 
Existing 3D human reconstruction approaches can be categorized into two main categories depending on whether explicit or implicit representations are used. Explicit representations mainly utilize the pre-defined mesh topology, such as statistical parametric models~\cite{loper2015smpl,xu2020ghum,osman2020star,anguelov2005scape} or personalized mesh registrations~\cite{gall2009motion,de2008performance}, to model naked human bodies with various poses and shapes.
To enhance the expressiveness, recent works have developed expressive statistical models capable of representing details beyond major human body~\cite{smplx, osman2022supr,feng2021collaborative} or introduced the surface deformation to capture fine-grained features~\cite{kolotouros2019convolutional,tang2019neural}. On the other hand, leveraging the remarkable advances in implicit neural representations~\cite{mescheder2019occupancy,mildenhall2020nerf}, another line of research has proposed to either rely purely on implicit representations~\cite{saito2019pifu} or combine it with statistical models~\cite{xiu2022icon,peng2021neural,chen2022gpnerf} to reconstruct expressive 3D human bodies. The most recent work ~\cite{dong2022totalselfscan,shen2023x} proposed to learn a single full-body avatar from multi-part portrait videos or 3D scans. 
In contrast, our approach focuses on developing 3D generative model for fully animatable human avatars, which is trainable on only unstructured 2D image collections.

\section{Method}
\begin{figure}[t]
\vspace{-1em}
\centering
\includegraphics[width=\textwidth]{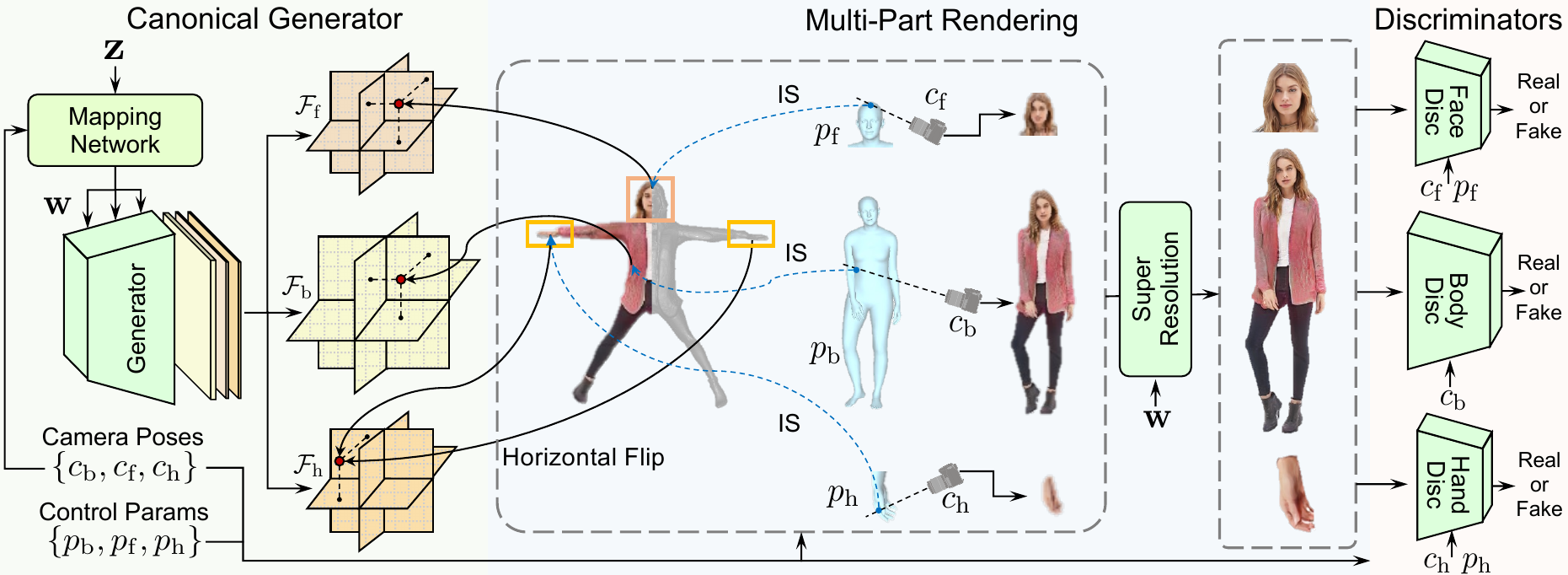}
\vspace{-3mm}
\caption{
Pipeline of \ours{}. Given a random noise \(\mathbf{z}\), the canonical generator synthesizes the avatar in the format of canonical multi-part and multi-scale Tri-planes given the corresponding camera pose \(c_\text{b}\). 
We then deform the canonical avatar under the guidance of control parameters \(p_\text{*}\) to render multi-part images using respective camera poses \(c_\text{*}\) and upsample the images using a super-resolution module. Discriminators encode the output images, camera poses, and control parameters into real or fake probabilities to critique the rendered images. IS represents inverse skinning.
}
\label{fig:network}
\end{figure}

In this section, we introduce \ours{}, a 3D generative model for synthesizing photo-realistic human avatars with expressive and disentangled controllability over facial expression, shape, jaw pose, body pose, and hand pose. Figure~\ref{fig:network} depicts the pipeline of our method. 

Given a random noise \(\mathbf{z}\) sampled from Gaussian distribution, \ours{} first synthesizes a human avatar with canonical body, face, and hand configurations. \textcolor{black}{In this work, we use X-pose~\cite{liu2021neural} and neutral shape, face, and hand as canonical configurations}. We leverage Tri-plane~\cite{chan2022efficient} as the fundamental building block of 3D representation in our canonical generator. To increase the capability of 3D representation for the smaller-scale face and hands, we introduce multi-part and multi-scale designs into the canonical Tri-plane (Sec.~\ref{sec:repr}). A mapping network first encodes \(\mathbf{z}\) and the camera viewpoint of body \(c_\text{b}\) into latent code \(\mathbf{w}\). The canonical generator then synthesizes three Tri-planes 
\(\mathcal{F}_k\) conditioned on \(\mathbf{w}\), where \(k\in\{\text{b},\text{f},\text{h}\}\) which stands for \(\{\text{body},\text{face},\text{hand}\}\). 

Based on the generated canonical avatar, we deform it from canonical space to observation space under the guidance of control signal \(p_\text{b}\) parameterized by an expressive statistical full body model, \ie, SMPL-X~\cite{smplx}. We adopt volumetric rendering~\cite{max1995optical} to synthesize the full body image. However, due to the scale imbalance between the face/hands and body, rendering only the full body image cannot guarantee quality for these detailed regions. To address this issue, we propose a multi-part rendering technique (Sec.~\ref{sec:render}). Specifically, we employ part-aware deformation and rendering based on the control parameters (\(p_\text{f}\) and \(p_\text{h}\)) and cameras (\(c_\text{f}\) and \(c_\text{h}\)). Accordingly, to ensure the plausibility and controllability of the generated avatars, we develop multi-part discriminators to critique the rendered images (Sec.~\ref{sec:disc}).

\subsection{Multi-scale and Multi-part Representation}
\label{sec:repr}
\ours{} is designed for expressive human avatars with an emphasis on the high-quality face and hands. However, the scale imbalance between face/hands and body may hamper the fidelity of the corresponding regions.
To address this issue, we propose a simple yet effective \textcolor{black}{multi-scale and multi-part representation} for expressive human avatar generation. Our multi-scale representation builds upon the efficient 3D representation, \ie, Tri-plane~\cite{chan2022efficient}, which stores the generated features on three orthogonal planes. Specifically, we design three Tri-planes for body, face, and hands, denoted as \(\mathcal{F}_\text{b}\in\mathbb{R}^{W_\text{b}\times W_\text{b}\times 3C}\), \(\mathcal{F}_\text{f}\in\mathbb{R}^{W_\text{f}\times W_\text{f}\times 3C}\), and \(\mathcal{F}_\text{h}\in\mathbb{R}^{W_\text{h}\times W_\text{h}\times 3C}\), respectively. The size of the face and hand Tri-planes is set to half of the body Tri-plane, with \(W_\text{f}=W_\text{h}=W_\text{b}/2\). 

As depicted in Figure~\ref{fig:network}, our canonical generator first synthesizes a compact feature map \(\mathcal{F}\in\mathbb{R}^{W_\text{b}\times W_\text{b}\times 9C/2}\), where \(C\) represents the number of channels. We then separate and reshape \(\mathcal{F}\) into \(\mathcal{F}_k\), where \(k\in\{\text{b},\text{f},\text{h}\}\), representing the canonical space of the generated human avatar. Furthermore, to save computation cost, we exploit the symmetry property of hands to represent both left and right hands using one single \(\mathcal{F}_\text{h}\) through a horizontal flip operation (refer to Appendix for details).

\subsection{Multi-part Rendering}
\label{sec:render}
Our method is trainable on unstructured 2D human images. Although this largely reduces the difficulty and cost to obtain data, the training is highly under-constrained due to the presence of diverse poses, faces, and clothes.
To facilitate the training process and improve the appearance quality, we propose a multi-part rendering strategy. This strategy allows \ours{} to learn each part based on the independent camera poses, which further enhances the geometry quality of the face and hands.


Specifically, for each training image, we utilize a pretrained model~\cite{feng2021collaborative} to estimate SMPL-X parameters \(\{p_\text{b},p_\text{f},p_\text{h}\}\) and camera poses \(\{c_\text{b},c_\text{f},c_\text{h}\}\) for body, face, and hands, respectively. In the rendering stage, we shoot rays using \(\{c_\text{b},c_\text{f},c_\text{h}\}\) and sample points \(\{\mathbf{x}_\mathbf{o}^\text{b}, \mathbf{x}_\mathbf{o}^\text{f}, \mathbf{x}_\mathbf{o}^\text{h}\}\) along the rays in the observation space. To compute the feature for each point, we employ inverse linear-blend skinning~\cite{lewis2000pose}, which finds the transformation of each point from observation space to canonical space produced by the canonical generator. Based on the parameter \(p_k\), where \(k\in\{\text{b},\text{f},\text{h}\}\), SMPL-X yields an expressive human body model \((\mathbf{v}, \mathbf{w})\), where \(\mathbf{v}\in\mathbb{R}^{N\times3}\) represents \(N\) vertices, and \(\mathbf{w}\in\mathbb{R}^{N\times J}\) represents the skinning weights of each vertex with respect to joint \(J\). For each point \(\mathbf{x}_\mathbf{o}^{k,i}\), where \(i=1\cdots M_k\) and \(M_k\) is the number of sampled points, we find its nearest neighbour \(\mathbf{n}\) from vertices \(\mathbf{v}\). We then compute the corresponding transformation from observation space to canonical space
\begin{equation}
    T^{k,i} = (\sum_j \mathbf{w}_j^{\mathbf{n}} \begin{bmatrix}
    R_j &t_j\\
    \boldsymbol{0}&\boldsymbol{1}
    \end{bmatrix}
    \begin{bmatrix}
    \boldsymbol{I} &\Delta^\mathbf{n}\\
    \boldsymbol{0}&\boldsymbol{1}
    \end{bmatrix}
    )^{-1},
\end{equation}
where \(j=1\cdots J\), \(R_j\) and \(t_j\) are derived from \(p_k\) with Rodrigues formula~\cite{bregler2004twist}, and \(\Delta^\mathbf{n}\) represents the offset caused by pose and shape for vertex \(\mathbf{n}\), which is calculated by SMPL-X. Based on this inverse transformation, we can calculate the coordinates for each point in canonical space \(\mathbf{x}_\mathbf{c}^{k,i}\) as
\begin{equation}
    \mathbf{x}_\mathbf{c}^{k,i} = T^{k,i} \mathbf{x}_\mathbf{o}^{k,i},
\end{equation}
where we apply homogeneous coordinates for the calculation. 

For the face and hands rendering, \ie, \(k \in \{\text{f},\text{h}\}\), we directly interpolate their corresponding Tri-plane \(\mathcal{F}_\text{f}\) and \(\mathcal{F}_\text{h}\) to compute the feature \(\mathbf{f}_\mathbf{c}^{\text{f},i}\) and \(\mathbf{f}_\mathbf{c}^{\text{h},i}\). Regarding the body rendering, we first define three bounding boxes \(\mathbb{B}_\text{f},\mathbb{B}_\text{lh},\mathbb{B}_\text{rh}\) for face, left and right hands in canonical body space. Then, we query canonical body points that are outside these bounding boxes from body Tri-plane \(\mathcal{F}_\text{b}\), while the canonical points inside these boxes 
from \(\mathcal{F}_\text{f}\) and \(\mathcal{F}_\text{h}\). The query process for body point \(\mathbf{x}_{\mathbf{c}}^{\text{b},i}\) is mathematically formulated as 
\begin{equation}
\mathbf{f}_\mathbf{c}^{\text{b},i}=\begin{cases}
Q(\mathbf{x}_\mathbf{c}^{\text{b},i}, \mathcal{F}_\text{f}), &\text{if}~\mathbf{x}_\mathbf{c}^{\text{b},i} \in \mathbb{B}_\text{f},\\
Q(\mathbf{x}_\mathbf{c}^{\text{b},i}, \mathcal{F}_\text{h}), &\text{if}~\mathbf{x}_\mathbf{c}^{\text{b},i} \in \{\mathbb{B}_\text{rh},\mathbb{B}_\text{lh}\},\\
Q(\mathbf{x}_\mathbf{c}^{\text{b},i}, \mathcal{F}_\text{b}),& \text{if}~\mathbf{x}_\mathbf{c}^{\text{b},i}\notin\{\mathbb{B}_\text{f},\mathbb{B}_\text{lh},\mathbb{B}_\text{rh}\},
\end{cases}
\end{equation}
where \(Q\) denotes querying the feature for the given point from the corresponding Tri-planes. 

Once the features \(\mathbf{f}_\mathbf{c}^{k,i}\) are obtained, they are encoded into color \(\mathbf{c}\) and geometry $d$ via two lightweight multi-layer perceptrons (MLP), where \(\mathbf{c}=\text{MLP}_\text{c}(\mathbf{f}_\mathbf{c}^{k,i})\). Inspired by prior works~\cite{or2022stylesdf,hong2023evad,zhang2023avatargen}, we employ signed distance field (SDF) as a proxy to model geometry. Additionally, following~\cite{hong2023evad,zhang2023avatargen}, we also query a base SDF \(d_\mathbf{c}\) in the canonical space, and predict delta SDF, such that \(d=d_\mathbf{c}+\text{MLP}_d(\mathbf{f}_\mathbf{c}^{k,i}, d_\mathbf{c})\). We then convert the SDF value into density \(\sigma=\frac{1}{\alpha} \text{Sigmoid}(\frac{-d}{\alpha})\) for volume rendering, where \(\alpha\) is a learnable parameter. 

To handle the body features queried from multiple Tri-planes, we apply feature composition on RGB and density using a window function~\cite{lombardi2021mixture} \textcolor{black}{for smoothness transition}. Specifically, if point \(\mathbf{x}_{\mathbf{c},\text{b}}^{k,i}\) is located in the overlapping region between the body and other parts (face, right hand, and left hand), their features are sampled from both Tri-planes and linearly blended together. More details on the feature composition can be found in the Appendix. Finally, volume rendering is applied to synthesize raw images for body, face, and hands, denoted as \(\{I_\text{b}^{\text{raw}}, I_\text{f}^{\text{raw}}, I_\text{h}^{\text{raw}}\}\). These raw images are then upsampled into high-resolution images \(\{I_\text{b}, I_\text{f}, I_\text{h}\}\) by a super-resolution module.

\subsection{Multi-part Discriminators}
\label{sec:disc}
Based on the images synthesized by \ours{} generator, we design a discriminator module to critique the generation results. 
To ensure both the fine-grained fidelity of appearance and geometry as well as disentangled control over the full body, including face and hands, we introduce multi-part discriminators to encode images \(\{I_\text{b}, I_\text{f}, I_\text{h}\}\) into real-fake scores for adversarial training. As depicted in Figure~\ref{fig:network}, these discriminators are conditioned on the respective camera poses to encode 3D priors, resulting in improved geometries as demonstrated in our experiments. To enhance the control ability of the face and hands, we further condition face discriminator on expression and shape parameters \([p_\text{f}^\psi, p_\text{f}^\beta]\), and condition hand discriminator on hand pose \(p_\text{h}^\theta\). 
We encode the camera pose and condition parameters into intermediate embeddings by two separate MLPs and pass them to the discriminators. The multi-part discriminator is formulated as
\begin{equation}
    s_k = \mathcal{D}_k(I_k, \text{MLP}_k^c(c_k)+ \text{MLP}_k^p(p'_k)), \text{where}~p'_k=\begin{cases}
        \varnothing, &\text{if}~k=\text{b}\\
        [p_\text{f}^\psi, p_\text{f}^\beta], &\text{if}~k=\text{f}\\
        p_\text{h}^\theta, &\text{if}~k=\text{h}
    \end{cases}.
\end{equation}
Here \(s_k\) denotes the probability of each image \(I_k\) being sampled from real data, and \(\mathcal{D}_k\) refers to the discriminator corresponding to the specific body part \(k\). \textcolor{black}{For body part, no conditioning parameters are used because we empirically find that the condition for body hinders the learning of appearance.}

\subsection{Training Losses}
The non-saturating GAN loss~\cite{goodfellow2020generative} is computed for each discriminator, resulting in \(L_\text{b}\), \(L_\text{f}\), and \(L_\text{h}\). We also regularize these discriminators using R1 regularization loss~\cite{mescheder2018training} \(L_\text{R1}\). To improve the plausibility and smoothness of geometry, we compute minimal surface loss \(L_\text{Minsurf}\), Eikonal loss \(L_\text{Eik}\), and human prior regularization loss \(L_\text{Prior}\) as suggested in previous works~\cite{or2022stylesdf,zhang2023avatargen}.

Due to the occlusion in the full body images, some training samples may not contain visible faces or hands. Thus, we balance the loss terms for both generator and discriminator based on the visibility of face \(\mathcal{M}_\text{f}\) and hands \(\mathcal{M}_\text{h}\), which denote whether face and hands are detected or not. The overall loss term of \ours{} is formulated as
\begin{equation}
    \begin{aligned}
L_\mathcal{G}=~L_\text{b}^\mathcal{G}+\lambda_\text{f}&\mathcal{M}_\text{f}\odot L_\text{f}^\mathcal{G} + \lambda_\text{h}\mathcal{M}_\text{h}^\mathcal{G}\odot L_\text{h}+\lambda_\text{Minsurf}L_\text{Minsurf}+\lambda_\text{Eik}L_\text{Eik}+\lambda_\text{Prior}L_\text{Prior},\\
    L_\mathcal{D}=~&L_\text{b}^\mathcal{D} + L_\text{R1}^\text{b}+\lambda_\text{f}\mathcal{M}_\text{f}\odot(L_\text{f}^\mathcal{D} + L_\text{R1}^\text{f})+\lambda_\text{h}\mathcal{M}_\text{h}\odot(L_\text{h}^\mathcal{D} + L_\text{R1}^\text{h}),
    \end{aligned}
\end{equation}
where \(\odot\) means instance-wise multiplication, and \(\lambda_*\) are the weighting factors for each term.

\section{Experiments}
We evaluate the performance of \ours{} on four datasets, \ie, DeepFashion~\cite{liu2016deepfashion}, MPV~\cite{zablotskaia2019dwnet}, UBC~\cite{dong2019towards}, and SHHQ~\cite{fu2022stylegan}. These datasets contain diverse full body images of clothed individuals. For each image in the dataset, we process it to obtain aligned body, face and hand crops, and their corresponding camera poses and SMPL-X parameters. Please refer to Appendix for more details. 
\subsection{Comparisons}
{\bf Baselines.}
We compare \ours{} with four state-of-the-art 3D GAN models for animatable human image generation: ENARF~\cite{noguchi2022unsupervised}, EVA3D~\cite{hong2023evad}, AvatarGen~\cite{zhang2023avatargen}, and AG3D~\cite{dong2023ag3d}. All these methods utilize 3D human priors to enable the controllability of body pose. ENARF conditions on sparse skeletons, while others condition on SMPL~\cite{loper2015smpl} model. Additionally, AvatarGen and AG3D incorporate an extra face discriminator to enhance face quality. We adopt the official implementations of ENARF and EVA3D, and cite results from AG3D directly. As for AvatarGen, it is reproduced and conditioned on SMPL-X to align with the setup of our model.

{\bf Quantitative comparisons.} The fidelity of synthesized image is measured by Frechet Inception Distance (FID)~\cite{heusel2017gans} computed between \(50K\) generated images and all the available real images in each dataset. To study the appearance quality for face and hands, we further crop face (resolution \(64^2\)) and hands (resolution \(48^2\)) regions from the generated and real images to compute \(\text{FID}_\text{f}\) and \(\text{FID}_\text{h}\). To evaluate pose control ability, we compute Percentage of Correct Keypoints (PCK) between \(5K\) real images and images generated using the same pose condition parameters of real images under a distance threshold of 0.1. To evaluate this ability in face and hand regions, we also report \(\text{PCK}_\text{f}\) and \(\text{PCK}_\text{h}\). Another critical evaluation for a fully controllable generative model is the disentangled control of fine-grained attributes. Inspired by previous works~\cite{deng2020disentangled,xu2023omniavatar}, we select one attribute from \(\{\text{expression}, \text{shape}, \text{jaw pose}, \text{body pose}, \text{hand pose}\}\), and modify the selected attribute while keeping others fixed for each synthesis. We then estimate the SMPL-X parameters for \(1K\) generated images \textcolor{black}{using a pre-trained 3D human reconstruction model~\cite{feng2021collaborative}} and compute the Mean Square Error (MSE) for the selected attribute between the input and estimated parameters. 

\begin{table}[h]
 \vspace{-1em}
 \renewcommand{\tabcolsep}{1.2pt}
 \small
  \caption{Quantitative comparisons with baselines in terms of appearance and overall control ability, with best results in \textbf{bold}. F.Ctl. indicates whether the approach generates fully controllable human body or not. \(^*\)We implement AvatarGen by conditioning it on SMPL-X.}
  \label{tab:comp1}
  \centering
  \begin{tabular}{lccccccccccccc}
    \toprule
    &&\multicolumn{6}{c}{DeepFashion~\cite{liu2016deepfashion}}           &\multicolumn{6}{c}{MPV~\cite{dong2019towards}}\\
    \cmidrule(r){3-8}\cmidrule(r){9-14}
             & F.Ctl. & FID\(\downarrow\)             & FID\(_\text{f}\)\(\downarrow\)  &FID\(_\text{h}\)\(\downarrow\) &PCK\(\uparrow\)  &PCK\(_\text{f}\)\(\uparrow\) &PCK\(_\text{h}\)\(\uparrow\) & FID\(\downarrow\)            & FID\(_\text{f}\)\(\downarrow\)  &FID\(_\text{h}\)\(\downarrow\) &PCK\(\uparrow\) &PCK\(_\text{f}\)\(\uparrow\) &PCK\(_\text{h}\)\(\uparrow\)\\
    \midrule
    ENARF~\cite{noguchi2022unsupervised} &\xmark &68.62 &52.17 &46.86 &3.54 &3.79 &1.34 &65.97 &47.71 &37.08 &3.06 & 3.55 &0.67\\
    EVA3D~\cite{hong2023evad} &\xmark &15.91  &14.63 &48.10 &56.36 &75.43 &23.14 &14.98 &27.48 &32.54 &33.00 &42.47 &19.24\\
    AG3D~\cite{dong2023ag3d} &\xmark &10.93 &14.79 &- &- &- &- &- &- &-&- &- &-\\ 
    AvatarGen~\cite{zhang2023avatargen}\(^*\) &\cmark &9.53 &13.96 &27.68 &60.12 &73.38 & 46.50 &10.06 &13.08 &19.75 &38.32 &45.26 &30.75\\
    \ours{} (Ours) &\cmark &{\bf 8.55} &{\bf 10.69} &{\bf 24.26} &{\bf 66.04} &{\bf 87.06} &{\bf 47.56} &{\bf 7.94} &{\bf 12.07} &{\bf 17.35} &{\bf 48.84} &{\bf 63.77} &{\bf 32.01}\\
    \midrule
    &&\multicolumn{6}{c}{UBC~\cite{zablotskaia2019dwnet}}  &\multicolumn{6}{c}{SHHQ~\cite{fu2022stylegan}}\\
    \cmidrule(r){3-8}\cmidrule(r){9-14}
             & F.Ctl. & FID\(\downarrow\)             & FID\(_\text{f}\)\(\downarrow\)  &FID\(_\text{h}\)\(\downarrow\) &PCK\(\uparrow\)  &PCK\(_\text{f}\)\(\uparrow\) &PCK\(_\text{h}\)\(\uparrow\) & FID\(\downarrow\)            & FID\(_\text{f}\)\(\downarrow\)  &FID\(_\text{h}\)\(\downarrow\) &PCK\(\uparrow\) &PCK\(_\text{f}\)\(\uparrow\) &PCK\(_\text{h}\)\(\uparrow\)\\
    \midrule
    ENARF~\cite{noguchi2022unsupervised} &\xmark &36.39 &34.27 &32.72 &6.90 &7.44 &6.37 &79.29 &50.19 &46.97 &4.43 & 4.62 &2.71\\
    EVA3D~\cite{hong2023evad} &\xmark &12.61  &36.87 &45.66 &36.31 &55.31 &8.38 &11.99 &20.04 &39.83 &31.24 &37.60 &18.38\\
    AG3D~\cite{dong2023ag3d} &\xmark &11.04 &15.83 &- &- &- &- &- &- &-&- &- &-\\ 
    AvatarGen~\cite{zhang2023avatargen}\(^*\) &\cmark &9.75 &13.23 &18.09 &65.31 &77.09 & 55.09 &10.52 &12.57 &28.21 &59.18 &78.71 &36.29\\
    \ours{} (Ours) &\cmark &{\bf 8.80} &{\bf 9.82} &{\bf 16.72} &{\bf 69.18} &{\bf 84.18} &{\bf 55.17} &{\bf 5.88} &{\bf 10.06} &{\bf 19.23} &{\bf 65.14} &{\bf 91.44} &{\bf 38.53}\\
    \bottomrule
  \end{tabular}
\end{table}

Table~\ref{tab:comp1} summarizes the results for appearance quality and pose control ability for body, face, and hands. It demonstrates that \ours{} outperforms existing methods \textit{w.r.t.} all the evaluation metrics, indicating its superior performance in generating controllable photo-realistic human images with high-quality face and hands. Notably, \ours{} shows significant improvements over the most recent method AG3D, achieving more than 20\% improvement in FID and \(\text{FID}_\text{f}\) on both DeepFashion and UBC datasets. Additionally, \ours{} achieves state-of-the-art pose control ability, with substantial performance boost in \(\text{PCK}_\text{f}\), \eg, a relative improvement of 40.90\% on MPV dataset against baseline.

Table~\ref{tab:comp2} presents the results for the disentangled control ability of \ours{} compared to the baseline methods. It is worth noting that ENARF and EVA3D are not fully controllable, but we still report all the evaluation metrics for these two methods to show the controllability lower bound. Notably, the generated images of ENARF are blurry. Thus, our pose estimator cannot estimate precise jaw poses, which leads to an outlier on UBC jaw pose. In general, \ours{} demonstrates state-of-the-art performance for fine-grained controls, \textcolor{black}{particularly in expression, jaw, and hand pose, improving upon baseline by 38.29\%, 25.93\%, and 33.87\% respectively on SHHQ dataset which contains diverse facial expressions and hand gestures}. These results highlight the effectiveness of \ours{} in enabling disentangled control over specific attributes of the generated human avatar images.

\begin{table}[h]
 \renewcommand{\tabcolsep}{2.2pt}
 \small
  \caption{Quantitative comparisons with baselines in terms of disentangled control ability measured by MSE. We report Jaw\(\times10^\mathrm{-4}\) and others \(\times10^\mathrm{-2}\) for simplicity, with {best results  in \textbf{bold}.} \(^*\)We implement AvatarGen by conditioning it on SMPL-X.}
  \label{tab:comp2}
  \centering
  \begin{tabular}{lcccccccccc}
    \toprule
    &\multicolumn{5}{c}{DeepFashion~\cite{liu2016deepfashion}}  &\multicolumn{5}{c}{MPV~\cite{dong2019towards}}\\
    \cmidrule(r){2-6}\cmidrule(r){7-11}
             &Exp\(\downarrow\) &Shape\(\downarrow\) &Jaw\(\downarrow\) &Body\(\downarrow\) &Hand\(\downarrow\) &Exp\(\downarrow\) &Shape\(\downarrow\) &Jaw\(\downarrow\) &Body\(\downarrow\) &Hand\(\downarrow\)\\
    \midrule
    ENARF~\cite{noguchi2022unsupervised} &13.47 &6.30 &5.79 &3.14 &9.87 &11.21 &4.91 &8.36 &2.75 &12.90\\
    EVA3D~\cite{hong2023evad} &6.03  &2.87 &5.11 &1.78 &3.68 &9.97 &4.14 &13.83 &1.80 &4.65\\
    AvatarGen~\cite{zhang2023avatargen}\(^*\) &4.92 &3.06 &5.05 &{\bf 1.23} &3.17 &8.98 &{\bf 3.88} &15.22 &1.11 &3.47\\ 
    \ours{} (Ours) &{\bf 4.46} &{\bf 2.77} &{\bf 3.67} &1.26 &{\bf 2.95} &{\bf 6.31} &{\bf 3.88} &{\bf 7.43} &{\bf 0.94} &{\bf 2.23}\\
    \midrule
    &\multicolumn{5}{c}{UBC~\cite{zablotskaia2019dwnet}}&\multicolumn{5}{c}{SHHQ~\cite{fu2022stylegan}}\\
    \cmidrule(r){2-6}\cmidrule(r){7-11}
             &Exp\(\downarrow\) &Shape\(\downarrow\) &Jaw\(\downarrow\) &Body\(\downarrow\) &Hand\(\downarrow\)&Exp\(\downarrow\) &Shape\(\downarrow\) &Jaw\(\downarrow\) &Body\(\downarrow\) &Hand\(\downarrow\)\\
    \midrule
    ENARF~\cite{noguchi2022unsupervised} &10.70 &6.11 &{\bf 3.62} &1.07 &8.19 &14.51 &6.43 &8.16 &3.27 &9.83\\
    EVA3D~\cite{hong2023evad} &7.00  &2.98 &5.36 &1.00 &2.78 &7.43 &4.15 &9.26 &1.93 &5.15\\
    AvatarGen~\cite{zhang2023avatargen}\(^*\) &9.59 &4.50 &9.34 &1.22 &3.01 & 9.01 &3.99 &8.87 &1.52 &4.99\\
    \ours{} (Ours) &{\bf 5.35} &{\bf 2.57} &4.76 &{\bf 0.73} &{\bf 1.63} &{\bf 5.56} &{\bf 3.66} &{\bf 6.57} &{\bf 1.24} &{\bf 3.30}\\
    \bottomrule
  \end{tabular}
\end{table}

{\bf Qualitative comparisons.}
\begin{figure}[t]
\vspace{-1em}
\centering
\includegraphics[width=\textwidth]{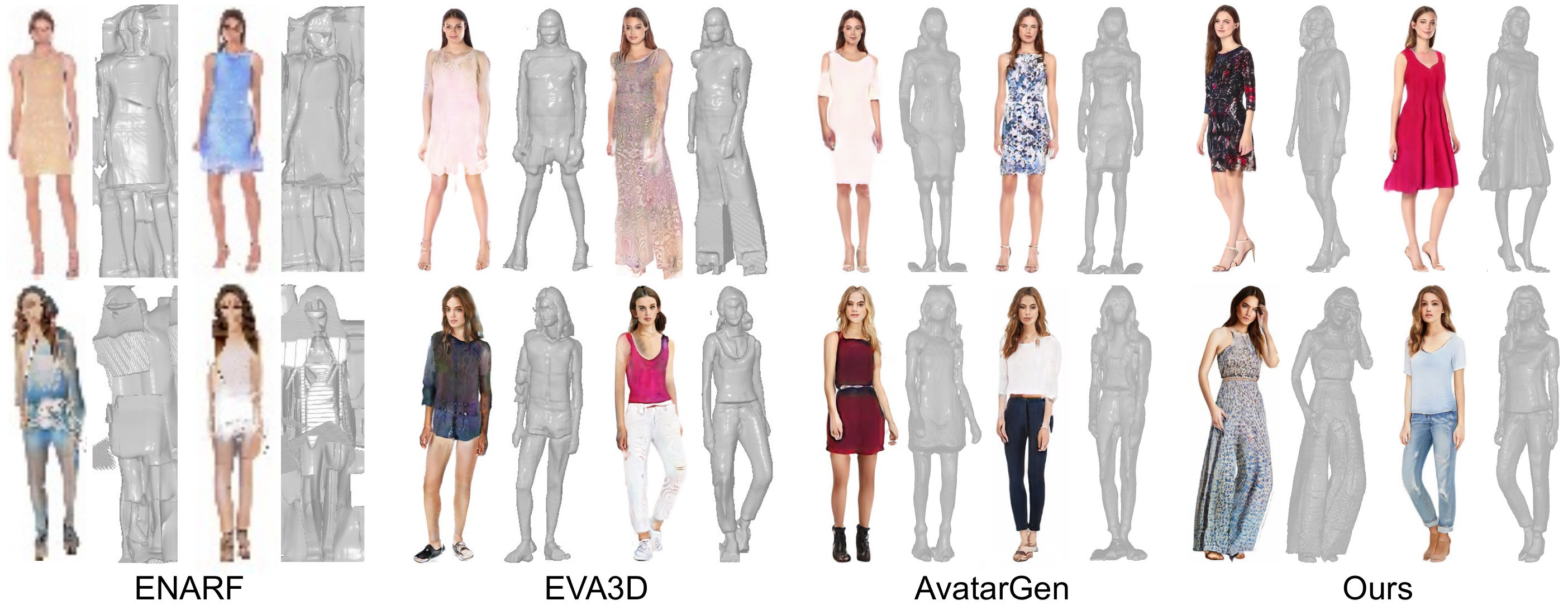}
\vspace{-1mm}
\caption{Comparisons against baselines in terms of appearance and 3D geometry. Our method produces photo-realistic human images with superior detailed geometries.}
\label{fig:qual_comp1}
\end{figure}
Figure~\ref{fig:qual_comp1} provides qualitative comparisons between \ours{} and baselines. From the results, we observe that ENARF struggles to produce reasonable geometry or realistic images due to the limitations of low training resolution. While EVA3D and AvatarGen achieve higher quality, they still fail to synthesize high-fidelity appearance and geometry for the face and hands. In contrast, \ours{} demonstrates superior performance with detailed geometries for face and hands regions, resulting in more visually appealing human avatar images. The improvement of \ours{} against baseline models is also confirmed by the perceptual user study, which is summarized in Table~\ref{tab:hstudy}. Notably, \ours{} achieves the best perceptual preference scores for both image appearance (\(\geq 57.2\%\)) and geometry (\(\geq 48.3\%\)) on all the benchmark datasets. 

\begin{figure}[t]
\vspace{-1em}
\centering
\includegraphics[width=\textwidth]{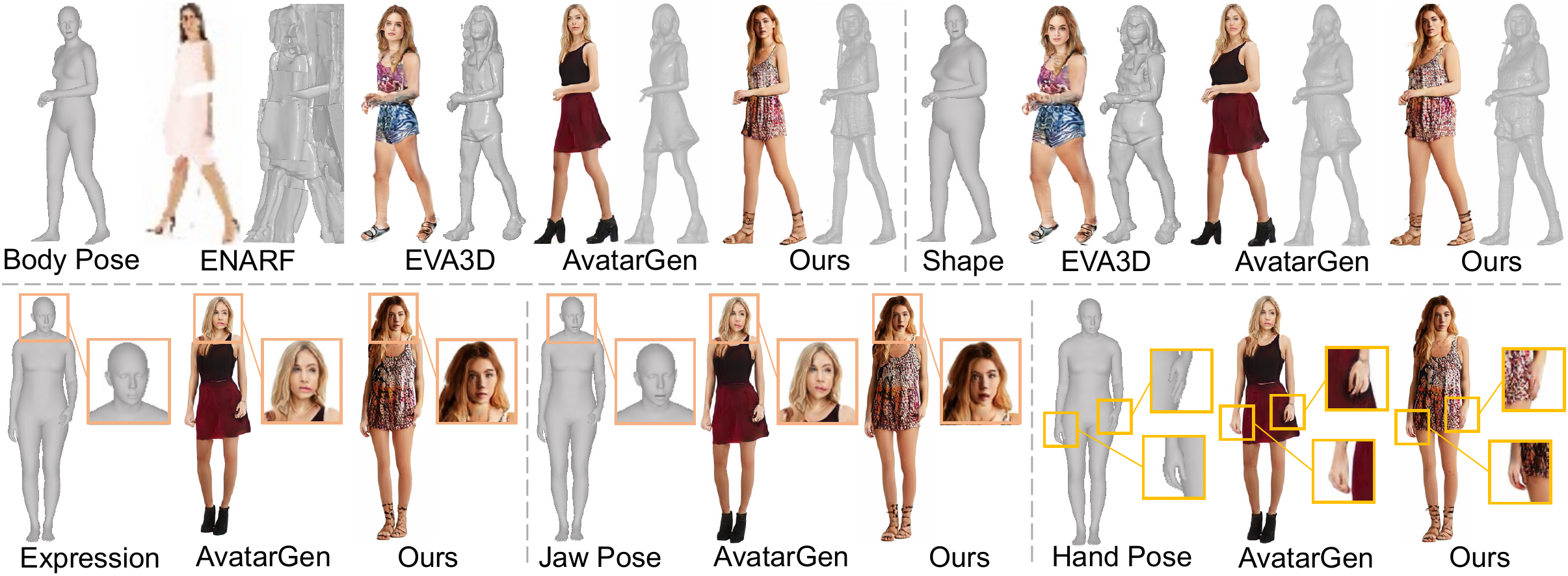}
\caption{Qualitative comparisons in terms of disentangled control ability. Our method exhibits state-of-the-art control abilities for body pose, shape, expression, jaw pose, and hand pose.}
\vspace{-4mm}
\label{fig:qual_comp2}
\end{figure}

Figure~\ref{fig:qual_comp2} showcases qualitative results for fine-grained control ability. We first observe that ENARF fails to generate a correct arm for the given body pose. Although EVA3D demonstrates a better pose condition ability, its shape conditioning ability is limited and the generated face suffers from unrealistic scaling.
On the other hand, AvatarGen shows comparable results for pose and shape control. However, when it comes to expression, jaw pose, and hand pose controls, ours significantly outperforms AvatarGen, \eg, AvatarGen produces distortion in mouth region and blurred fingers while \ours{} demonstrates natural faces and correct hand poses.

\begin{table}[t]
 \renewcommand{\tabcolsep}{2.2pt}
 \small
  \caption{We conduct a perceptual human study and report participants' preferences on images and geometries generated by our method and baselines. It is measured by preference rate (\%), with best results in \textbf{bold}. {\it RGB} represents image, and {\it Geo} represents geometry.}
  \label{tab:hstudy}
  \centering
  \begin{tabular}{lcccccccccc}
    \toprule
    &\multicolumn{2}{c}{DeepFashion}  &\multicolumn{2}{c}{MPV}&\multicolumn{2}{c}{UBC}&\multicolumn{2}{c}{SHHQ}\\
    \cmidrule(r){2-3}\cmidrule(r){4-5}\cmidrule(r){6-7}\cmidrule(r){8-9}
             &{\it RGB}\(\uparrow\) &{\it Geo}\(\uparrow\) &{\it RGB}\(\uparrow\) &{\it Geo}\(\uparrow\)&{\it RGB}\(\uparrow\) &{\it Geo}&{\it RGB}\(\uparrow\) &{\it Geo}\\
    \midrule
    ENARF&0.0 &0.0 &0.0 &0.0 &0.6 &0.0 &0.0 &0.0\\
    EVA3D &17.3  &35.6 &15.0 &17.2 &7.8 &34.4 &11.3 &15.5\\
    AvatarGen &15.4 &16.1 &17.2 &18.9 &34.4 &3.9 &28.2&28.6\\ 
    \ours{} (Ours) &{\bf 67.3} &{\bf 48.3} &{\bf 67.8} &{\bf 63.9} &{\bf 57.2} &{\bf 61.7} &{\bf 60.5} &{\bf 55.9}\\
    \bottomrule
  \end{tabular}
\end{table}

\begin{table}[t]
  \caption{Ablations of our method on SHHQ dataset. We vary our representation, rendering method, and discriminators to investigate their effectiveness.}
  \label{tab:ab}
  \renewcommand{\tabcolsep}{1.8pt}
  \small
\begin{subtable}[!t]{0.3\linewidth}
    \centering
  \begin{tabular}{cccc}
    \toprule
    {\it Repr.}     & FID\(\downarrow\)      &FID\(_\text{f}\)\(\downarrow\)  &FID\(_\text{h}\)\(\downarrow\)\\
    \midrule
    w/o both &11.50 &12.57 &20.97 \\
    w/ face &11.27 &11.95 &20.10 \\
    w/ hand &9.64 &11.61 &19.92 \\
    w/ both &5.88 &10.06 &19.23 \\
    \bottomrule
  \end{tabular}
  \caption{The effect of multi-scale and multi-part representations.}
  \label{tab:ab_repr}
\end{subtable}
\hspace{\fill}
\begin{subtable}[!t]{0.35\linewidth}
    \centering
  \begin{tabular}{ccccc}
    \toprule
    {\it Face}     & FID\(\downarrow\)      &FID\(_\text{f}\)\(\downarrow\)  &Exp\(\downarrow\)&Jaw\(\downarrow\)\\
    \midrule
    w/o Rend &14.53 &20.63 &6.58 &7.26\\
    w/o Disc &7.40 &9.20 &6.27 &6.58\\
    w/ both &5.88 &10.06 &5.56 &6.57\\
    \bottomrule
  \end{tabular}
  \caption{The effect of face rendering and face discriminator.}
  \label{tab:ab_face}
\end{subtable}
\hspace{\fill}
\begin{subtable}[!t]{0.3\linewidth}
    \centering
  \begin{tabular}{ccccc}
    \toprule
    {\it Hand}     & FID\(\downarrow\)      &FID\(_\text{h}\)\(\downarrow\)  &Hand\(\downarrow\)\\
    \midrule
    w/o Rend &14.28 &26.66 &4.51 \\
    w/o Disc &7.78 &16.74 &4.46 \\
    w/ both &5.88 &19.23 &3.33 \\
    \bottomrule
  \end{tabular}
  \caption{The effect of hand rendering and hand discriminator.}
  \label{tab:ab_hand}
\end{subtable}
\vspace{-3mm}
\end{table}

\subsection{Ablation studies}
To verify the design choices in our method, we conduct ablation studies on SHHQ dataset, which contains diverse appearances, \ie, various human body, face, and hand poses as well as clothes. 

\textbf{Representation.} \ours{} adopts a multi-scale and multi-part representation to improve the quality for face and hands regions. We study the necessity of this design by removing Tri-planes for face and hands. Table~\ref{tab:ab_repr} provides the results, indicating that using only a single full-body Tri-plane (without any specific Tri-planes for face or hands) results in a significant degradation in appearance quality. Adding either face or hand Tri-plane can alleviate this issue and all the FID metrics drop slightly. The best results are achieved when both face and hand Tri-planes are enabled, demonstrating the importance of our multi-scale and multi-part representation.

\begin{figure}[t]
\centering
\includegraphics[width=0.95\textwidth]{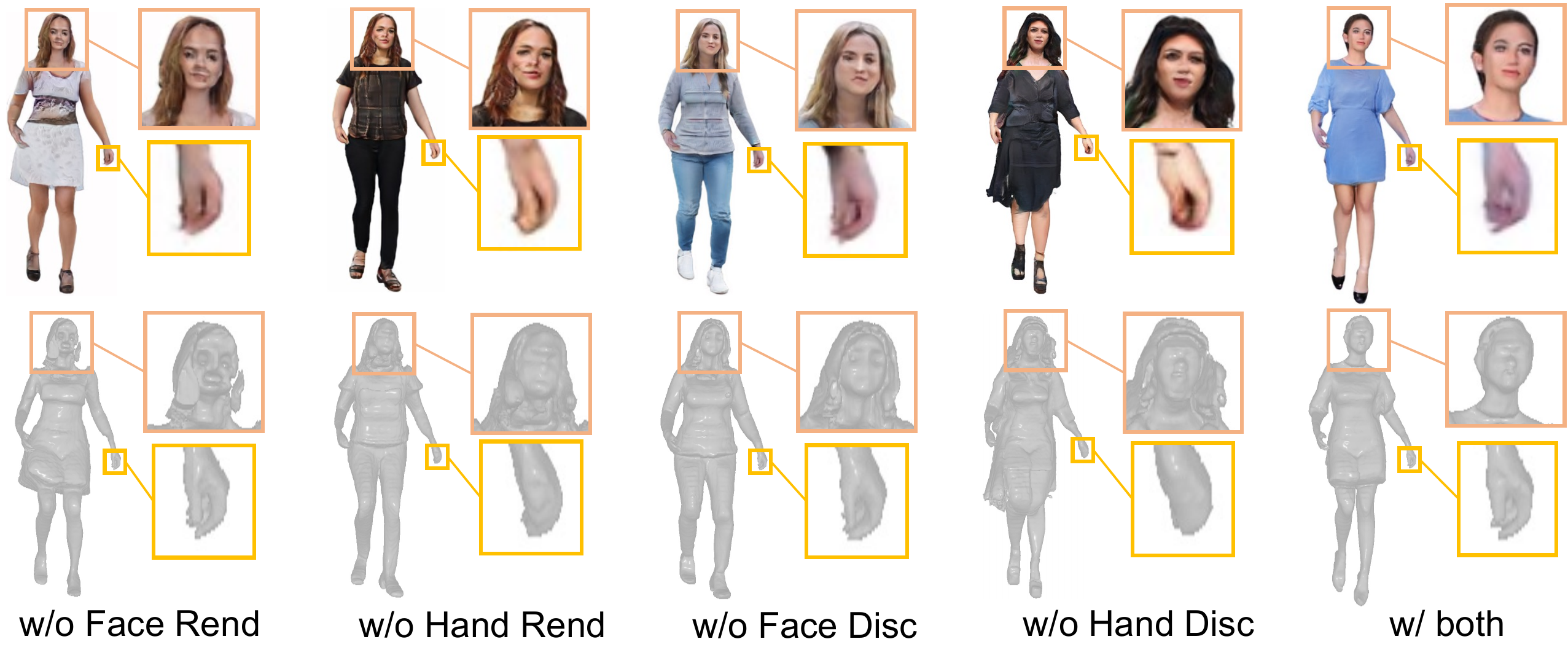}
\caption{Qualitative results for the ablations on multi-part rendering and discriminators.}
\vspace{-0.8em}
\label{fig:ab}
\end{figure}

\textbf{Multi-part rendering.} In our model, we render multiple parts independently in the forward process to disentangle the learning of body, face, and hands. Table~\ref{tab:ab_face} demonstrates that independent rendering for face is crucial, as it significantly improves both fidelity (FID\(_\text{f}\): 20.63 {\it vs.} 10.06) and control ability  (Exp: 6.58 {\it vs.} 5.56, Jaw: 7.26 {\it vs.} 6.57) for face. Similarly, without rendering for hand, FID\(_\text{h}\) increases from 18.85 to 25.94, and MSE increases from 3.28 to 4.55 (Table~\ref{tab:ab_hand}). 
The effectiveness of multi-part rendering is further supported by the qualitative results shown in Figure~\ref{fig:ab}. Without independent rendering, the geometry quality degrades. For example, the eyes and mouth are collapsed without face rendering, and the model also fails to synthesize geometric details for hand when hand rendering is disabled. These highlight the importance of multi-part rendering in facilitating the learning of 3D geometries for different body parts.

\textbf{Discriminators.} To study the effect of multi-part discriminators, we disable each of them during training. 
As shown in Table~\ref{tab:ab_face},
without face discriminator, the overall appearance quality deteriorates. 
Despite the slight improvement in face appearance,
there is a drop in the control ability, as evidenced by the increase in the MSE values for expression and jaw pose. A similar observation can be made for hand discriminator in Table~\ref{tab:ab_hand}. Furthermore, the qualitative results shown in Figure~\ref{fig:ab} provide visual evidence of the impact of the face and hand discriminators on the 3D geometries. When they are removed, the geometries for face and hand collapse. 

\subsection{Applications}
\begin{figure}[t]
     \centering
     \begin{subfigure}[b]{0.44\linewidth}
         \centering
         \includegraphics[width=\textwidth]{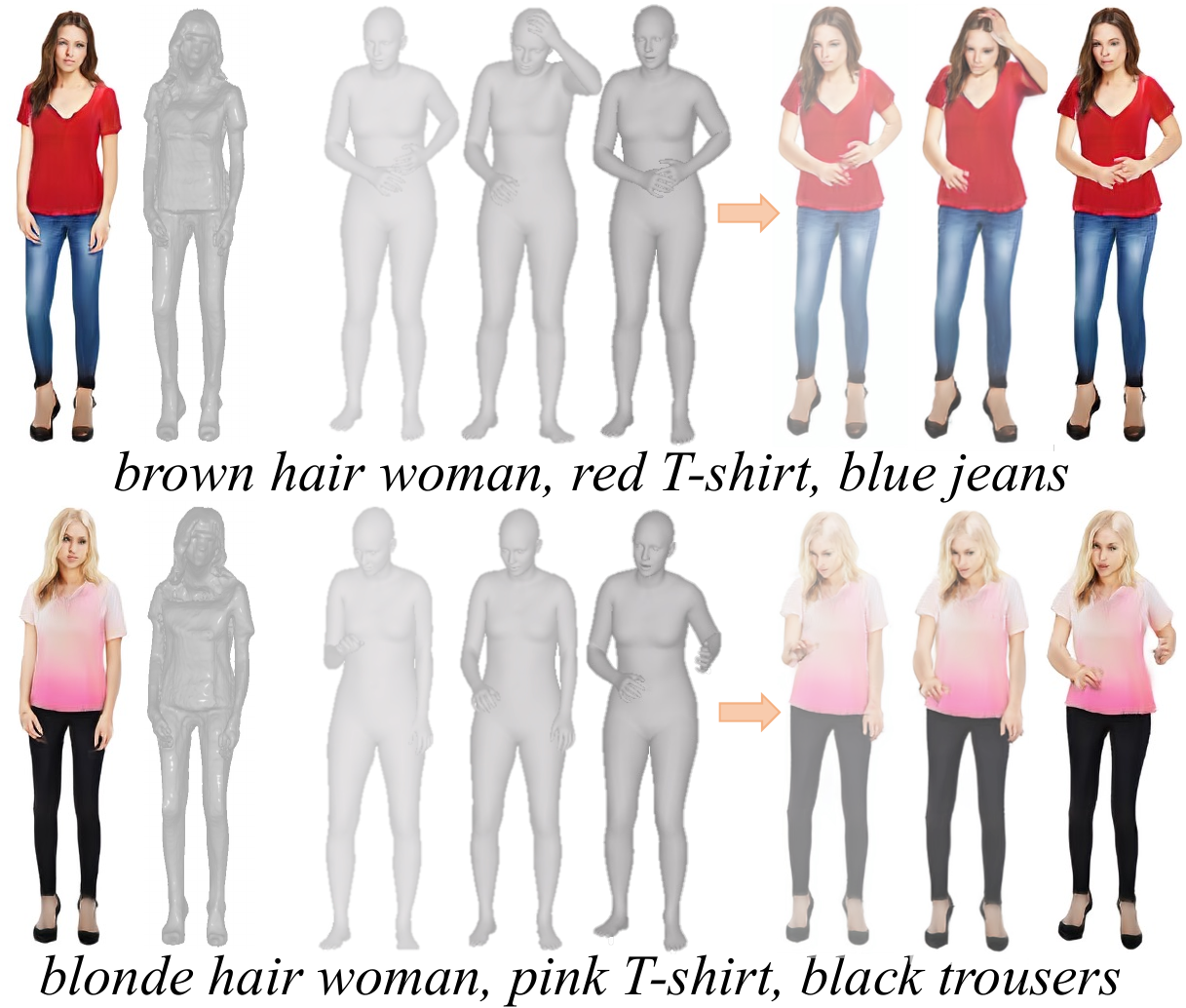}
         \caption{Text-guided avatar synthesis.}
         \label{fig:app:text}
     \end{subfigure}
     \hfill
     \begin{subfigure}[b]{0.55\linewidth}
         \centering
         \includegraphics[width=\textwidth]{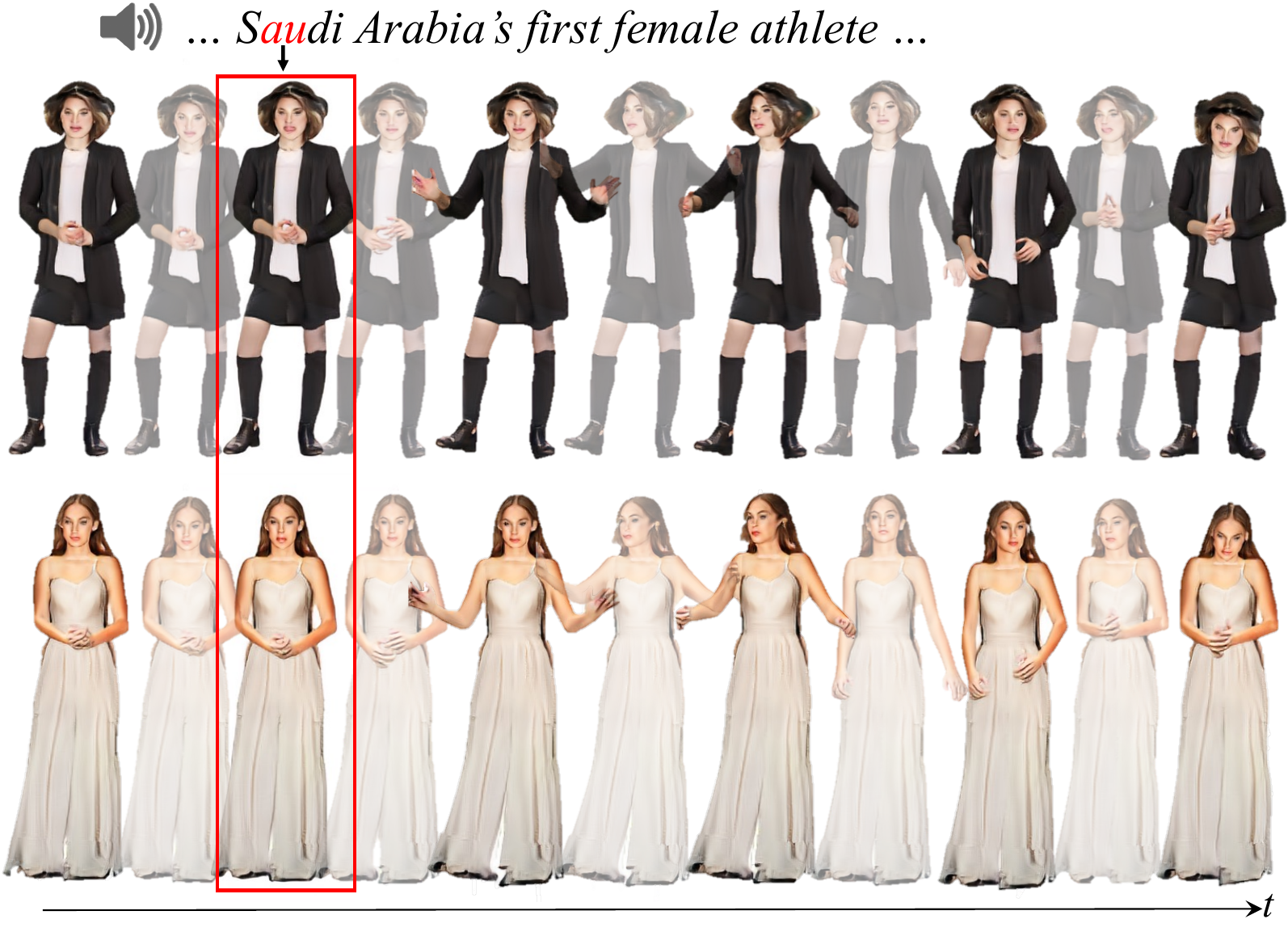}
         \caption{Audio-driven avatar animation.}
         \label{fig:app:talk}
     \end{subfigure}
        \caption{Downstream applications of our method.}
        \label{fig:app}
        \vspace{-1em}
\end{figure}
\textbf{Text-guided avatar synthesis.} Inspired by recent works~\cite{hong2022avatarclip,zhang2023avatargen,youwang2022clip} on text-guided avatar generation, we leverage a pretrained vision-language encoder CLIP~\cite{radford2021learning} to guide the generation process using the given text prompt. The text-guided avatar generation process involves randomly sampling a latent code \(\mathbf{z}\) and a control parameter \(p_\text{b}\) from the dataset,
and optimizing \(\mathbf{z}\) by maximizing the CLIP similarities between the synthesized image and text prompt. As shown in Figure~\ref{fig:app:text}, the generated human avatars exhibit the text-specified attributes, \ie, hair and clothes adhere to the given text prompt (\eg, brown hair and red T-shirt). The generated avatar can be re-targeted by novel SMPL-X parameters, allowing for additional control and customization of the synthesis.

\textbf{Audio-driven animation.} The ability of \ours{} to generate fully animatable human avatars with fine-grained control (Figure~\ref{fig:teaser}) opens up possibilities for audio-driven animation. The 3D avatars can be driven by arbitrary SMPL-X motion sequences generated by recent works such as \cite{yi2022generating} given audio inputs. Specifically, we sample an audio stream and SMPL-X sequence from TalkSHOW~\cite{yi2022generating} and use it to animate the generated avatars. As shown in Figure~\ref{fig:app:talk}, \ours{} is able to synthesize temporally consistent video animations where the jaw poses of the avatars are synchronized with the audio stream (highlighted in red box). Additionally, the generated avatars are generalizable given novel body poses and hand gestures, allowing diverse and expressive animations.

\section{Limitations}
Although \ours{} is able to synthesize photo-realistic and fully animatable human avatars, there are still areas where improvements can be made: 
(1) \ours{} relies on pre-estimated SMPL-X parameters, the inaccurate SMPL-X may introduce potential errors into our model, which can lead to artifacts and degraded body images. Please refer to {\it Sup. Mat.} for the experimental analysis of this issue. We believe our method can benefit from a more accurate SMPL-X estimation method or corrective operations. 
(2) SMPL-X only represents naked body. Thus, methods built upon SMPL-X could struggle with modeling loose clothing, which is a long-standing challenge for 3D human modeling. We believe an advanced human body prior or independent clothing modeling approach is helpful to alleviate this issue.
(3) Face and hand images in existing human body datasets lack diversity and sharpness, which affects the fidelity of our generation results, particularly for the novel hand poses that are out-of-distribution. A more diverse dataset with high-quality face and hand images could help tackle this problem. 
(4) \ours{} utilizes inverse blend skinning to deform the points from canonical space to the observation space. However, this process could introduce errors, particularly when computing nearest neighbors for query points located in the connection or interaction regions. Thus, exploring more robust and accurate techniques, such as forward skinning~\cite{chen2021snarf}, could open up new directions for future work.

\section{Conclusion}
This work introduces \ours{}, a novel 3D avatar generation framework that offers expressive control over facial expression, shape, body pose, jaw pose, and hand pose. Through the use of multi-scale and multi-part representation, \ours{} can model details for small-scale regions like faces and hands. By adopting multi-part rendering, \ours{} disentangles the learning process and produces realistic details for appearance and geometry. With multi-part discriminators, our model is capable of synthesizing high-quality human avatars with disentangled fine-grained control ability. The capabilities of \ours{} open up a range of possibilities for downstream applications, such as text-guided avatar synthesis and audio-driven animation.



\section*{Acknowledgement}
This project is supported by the National Research Foundation, Singapore under its NRFF Award NRF-NRFF13-2021-0008, and the Ministry of Education, Singapore, under the Academic Research Fund Tier 1 (FY2022).

\bibliography{Reference}


\end{document}


\maketitle

\section{Appendix}
\subsection{Feature Sampling}
We provide additional details on the feature query process for face, hand, and body.

\textbf{Query face and hand features.}
As described in the main paper, we directly query features for face and hand from their respective Tri-plane \(\mathcal{F}_\text{f}\) and \(\mathcal{F}_\text{h}\) to render their images. This process can be formulated as
\begin{equation}
\mathbf{f}_\mathbf{c}^{k,i}=Q(\mathbf{x}_\mathbf{c}^{k,i}, \mathcal{F}_{k}),
\end{equation}
where \(k \in \{\text{f}, \text{h}\}\). Here, \(Q\) denotes querying the feature from Tri-plane. Specifically, this process involves grid sampling (interpolation) operation (represented as \(\mathrm{Inter}\)) for each query point \(\mathbf{x}_\mathbf{c}^{k,i}\) in three orthogonal feature planes \(\{\mathcal{F}_{k}^X,\mathcal{F}_{k}^Y,\mathcal{F}_{k}^Z\}\). Prior to the query, we normalize the coordinates of the points using the bounding boxes \(\mathbb{B}_k\) defined in the canonical space for the face and hands. This normalization is performed as 
\begin{equation}
    \hat{\mathbf{x}}_\mathbf{c}^{k,i} = \frac{2\mathbf{x}_\mathbf{c}^{k,i}-(\mathbb{B}_k^{min}+\mathbb{B}_k^{max})}{\mathbb{B}_k^{max}-\mathbb{B}_k^{min}},
\end{equation}
Here, we use \(k \in \{\text{f}, \text{rh}\}\), indicating that only the right hand image is rendered in multi-part rendering process.
Consequently, the query process can be mathematically expressed as
\begin{equation}
Q(\mathbf{x}_\mathbf{c}^{k,i}, \mathcal{F}_{k}) = \sum_{t\in\{X,Y,Z\}} \mathrm{Inter}(\hat{\mathbf{x}}_\mathbf{c}^{k,i}, \mathcal{F}_{k}^t),
\end{equation}

\textbf{Query body features.} We apply a similar process to normalize the coordinates of body points and sample them from their respective Tri-planes based on which part they belong to.  In addition, during body image rendering, it is necessary to render both the left and right hands. To reduce computational cost, we exploit the symmetry between the left and right hands. Hence, we utilize a single hand Tri-plane, denoted as \(\mathcal{F}_\text{h}\) to model both hands simultaneously. Here, we explicitly define the bounding box \(\mathbb{B}_\text{lh}\) for left hand. For the body point falling within \(\mathbb{B}_\text{lh}\), we query their features by flipping the \(x\)-axis. Suppose a normalized left hand point is \(\hat{\mathbf{x}}_\mathbf{c}^{\text{lh},i}=[x, y, z]^T\), the flipped coordinate would be \(\hat{\mathbf{x}}{'}_\mathbf{c}^{\text{lh},i}=[-x, y, z]^T\). The query process for the left hand can be expressed as
\begin{equation}
Q(\mathbf{x}_\mathbf{c}^{\text{lh},i}, \mathcal{F}_{k}) = \sum_{t\in\{X,Y,Z\}}\mathrm{Inter}(\hat{\mathbf{x}}{'}_\mathbf{c}^{\text{lh},i}, \mathcal{F}_\text{h}^t),
\end{equation}

\begin{figure}[t]
\centering
\includegraphics[width=0.4\textwidth]{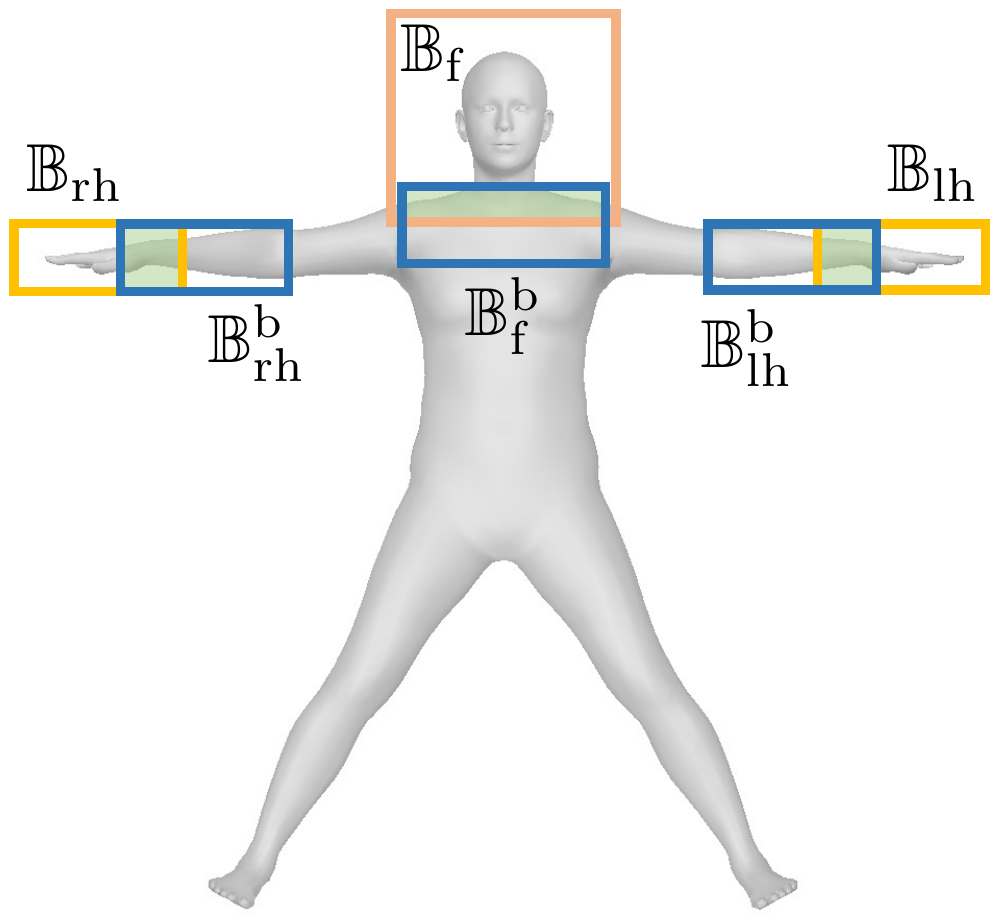}
\caption{We define the bounding boxes (\(\mathbb{B}_\text{f},\mathbb{B}_\text{lh},\mathbb{B}_\text{rh}\))in the canonical space. Overlapping regions (\(\{\mathbb{B}_\text{f}\cap\mathbb{B}_\text{f}^\text{b},\mathbb{B}_\text{rh}\cap\mathbb{B}_\text{rh}^\text{b},\mathbb{B}_\text{lh}\cap\mathbb{B}_\text{lh}^\text{b}\}\)) are highlighted in green.}
\label{fig:bbox}
\vspace{-3mm}
\end{figure}

\textbf{Composition of body features.}
As the body features are sampled from multi-part representations for the face and hand, there may be noticeable transitions in the overlapping regions between the body and other parts. To address this issue and enhance the smoothness of the transitions, we utilize a window function~\cite{lombardi2021mixture} to composite the features. As depicted in Figure~\ref{fig:bbox}, in addition to the bounding boxes for face and hands \(\mathbb{B}_\text{f},\mathbb{B}_\text{lh},\mathbb{B}_\text{rh}\), we further define three bounding boxes \(\mathbb{B}_\text{f}^\text{b},\mathbb{B}_\text{lh}^\text{b},\mathbb{B}_\text{rh}^\text{b}\) in the canonical space for the transition regions. For each point located in the overlapping regions \(\{\mathbb{B}_\text{f}\cap\mathbb{B}_\text{f}^\text{b},\mathbb{B}_\text{rh}\cap\mathbb{B}_\text{rh}^\text{b},\mathbb{B}_\text{lh}\cap\mathbb{B}_\text{lh}^\text{b}\}\) (highlight in green), we normalize their coordinates based on the bounding box of each part (\(\hat{\mathbf{x}}{'}_\mathbf{c}^{k,i}=[x, y, z]^T\)) and sample their features twice. The first sampling is performed from the body Tri-plane \(\mathcal{F}_\text{b}\), and the second sampling is performed from the corresponding part Tri-plane \(\mathcal{F}_k\). Subsequently, we encode the features into color \(\mathbf{c}\) and geometry $d$. The process of feature composition can be formulated as follows
\begin{equation}
\{\mathbf{c}, d\} = \frac{1}{\sum \omega} \sum \omega \{\mathbf{c}^k, d^k\},~\text{where}~\omega=\mathrm{exp}({-m(x^n + y^n + z^n)}), k\in\{\text{f}, \text{rh}, \text{lh}\}.
\end{equation}
Here, \(m\) and \(n\) are empirically chosen parameters, with \(m=2\), \(n=6\) in this work.

\subsection{Loss Terms}
{\bf Minimal surface loss.} Inspired by previous studies \cite{zhang2023avatargen,or2022styleSDF,hong2023evad}, we utilize the minimal surface loss to discourage the presence of spurious and invisible surfaces within the generated scene. This loss term guides the generator to create a human avatar with minimal volume of zero-crossings. Specifically, the SDF values that are in proximity to zero will be penalized by the loss term, which encourages a smoother and more coherent surface. This process is formulated as 
\begin{equation}
    L_\text{Minsurf} = \sum_{k,i} \mathrm{exp}({-100|d^{k,i}|}),
\end{equation}

{\bf Eikonal loss.} The Eikonal term is derived from the Eikonal equation, which ensures that the SDF defines a smooth boundary by enforcing its differentiability everywhere, specifically \(||\nabla d^{k,i}||=1\). Consequently, the Eikonal loss is defined as
\begin{equation}
    L_\text{Eik} = \sum_{k,i} (||\nabla d^{k,i}||-1).
\end{equation}

{\bf SMPL-X prior loss.} Despite representing a naked human body, the coarse geometric information encoded in the SMPL-X parametric model can still provide valuable guidance for training our avatar generator. Hence, we incorporate a regularization term to align the predicted geometry value \(d\) with the corresponding SDF value \(d_\mathbf{c}\) queried from the canonical SPML-X space. This loss term is expressed as
\begin{equation}
    L_\text{Prior} = \frac{1}{|\mathcal{R}|} \sum_{\mathbf{x}_\mathbf{c}^{k,i}\in \mathcal{R}}w ||(d^{k,i}-d_\mathbf{c}^{k,i})||, \text{where}~w=\mathrm{exp}\left(\frac{-(d_\mathbf{c}^{k,i})^{2}}{\kappa}\right),
\end{equation}
where \(\mathcal{R}\) represents the set of sampled rays.

\subsection{Inference}
\ours{} synthesizes the canonical avatar and renders body, face, and hand images using their respective cameras during training. However, in the inference stage, we exclude the part-aware rendering for face and hands. This means that we only render images for the full bodies as \ours{} focuses on human avatar generation. Even though only body image rendering is enabled during inference, we continue to utilize the same multi-scale and multi-part 3D representation. This canonical avatar representation still preserves the fine details for the face and hands and provides control ability for these regions.

\subsection{Implementation Details}
\label{appendix:impl}
We implement our model using PyTorch~\cite{paszke2019pytorch}, and optimized it using the Adam optimizer. The learning rate for the generator is set to \(2.5\times 10^{-3}\), while for the discriminator it is set to \(2\times 10^{-3}\). During the training stage, we employ a volume rendering resolution of \(224\times112\) for the full body, and \(28^2\) for the face and hands. Additionally, the super-resolution module upsamples body image into \(512\times 256\), face image into \(256^2\), and hand image into \(256^2\). For training, we utilize a depth resolution of \(40\), with \(20\) for hierarchical sampling, while we use a depth resolution of \(64\) in inference stage. The weighting factors for the loss terms are set as follows: \(\lambda_\text{f}=0.25\), \(\lambda_\text{h}=0.75\), \(\lambda_\text{Minsurf}=5\times 10^{-3}\), \(\lambda_\text{Eik}=1\times 10^{-3}\), and \(\lambda_\text{Prior}=1.0\). We apply R1 regularization with a R1 gamma value of \(10\). Our model is trained on 8 Nvidia V100 GPUs for 106 hours with a batch size of 16.

\subsection{Quantitative Results for Geometry}
In order to quantitatively evaluate the geometry quality of \ours{}, we employ depth metrics as suggested in a recent work~\cite{zhang2023avatargen}. We generate \(1K\) images and adopt a pretrained human digitalization model~\cite{saito2020pifuhd} to estimate the pseudo ground truth depth map for the synthesized images. We then calculate the Mean Square Error (MSE) between the pseudo labels and the depth maps rendered by volume rendering to assess the geometric quality. We compare \ours{} with the strongest baseline, AvatarGen, on four benchmark datasets. We report Depth following AvatarGen (at a resolution of \(128^2\)) and we also report Depth256 which is calculated at a higher resolution of \(256^2\) to evaluate the details of the geometry.
\begin{table}[h]
 \renewcommand{\tabcolsep}{3pt}
 \small
  \caption{Quantitative comparisons of geometric quality between \ours{} and AvatarGen on four benchmark datasets, with {best results in \textbf{bold}.}}
  \label{tab:comp3}
  \centering
  \begin{tabular}{lcccccccc}
    \toprule
    &\multicolumn{2}{c}{DeepFashion~\cite{liu2016deepfashion}}  &\multicolumn{2}{c}{MPV~\cite{dong2019towards}} &\multicolumn{2}{c}{UBC~\cite{zablotskaia2019dwnet}}&\multicolumn{2}{c}{SHHQ~\cite{fu2022stylegan}}\\
    \cmidrule(r){2-3}\cmidrule(r){4-5}\cmidrule(r){6-7}\cmidrule(r){8-9}
             &Depth\(\downarrow\) &Depth256\(\downarrow\)&Depth\(\downarrow\) &Depth256\(\downarrow\)&Depth\(\downarrow\) &Depth256\(\downarrow\)&Depth\(\downarrow\) &Depth256\(\downarrow\) \\
    \midrule
    AvatarGen~\cite{zhang2023avatargen} &.794 &.950 &.688 &.867&.806 &.958&.826 &.932\\ 
    \ours{} (Ours) &{\bf .474} &{\bf .597} &{\bf .527} &{\bf .651}&{\bf .482} &{\bf .613}&{\bf .438} &{\bf .560}\\
    \bottomrule
  \end{tabular}
\end{table}

Table~\ref{tab:comp3} summarizes the results for Depth metrics. \ours{} consistently outperforms the baseline method in terms of depth consistency for both Depth and Depth256 metrics across all datasets. It demonstrates that our method achieves a superior geometric quality, \eg, a relative improvement of 47.0\% for Depth and 39.9\% for Depth256 on SHHQ dataset.

\subsection{Additional Qualitative Results}
\begin{figure}[t]
\centering
\includegraphics[width=\textwidth]{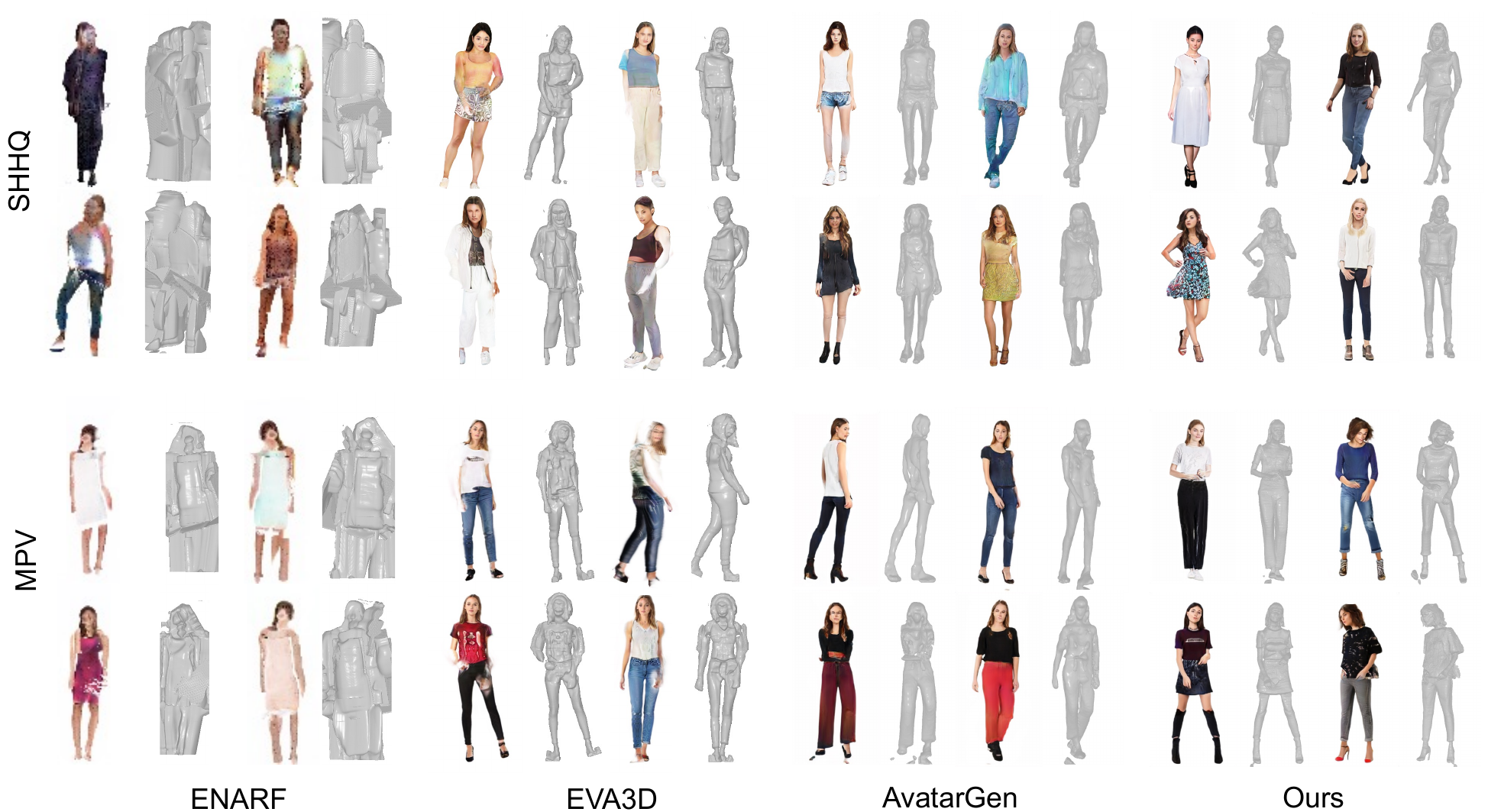}
\caption{Qualitative comparisons between \ours{} and baselines on SHHQ and MPV datasets. Best view in \(2\times\) zoom.}
\label{fig:qual2}
\vspace{-3mm}
\end{figure}

{\bf Qualitative comparisons.} We visualize more results in Figure~\ref{fig:qual2} for qualitative comparisons on MPV and SHHQ datasets. \ours{} outperforms baseline methods in terms of both visual quality and geometric details, which is also confirmed by our perceptual human study.

\begin{figure}[t]
\centering
\includegraphics[width=\textwidth]{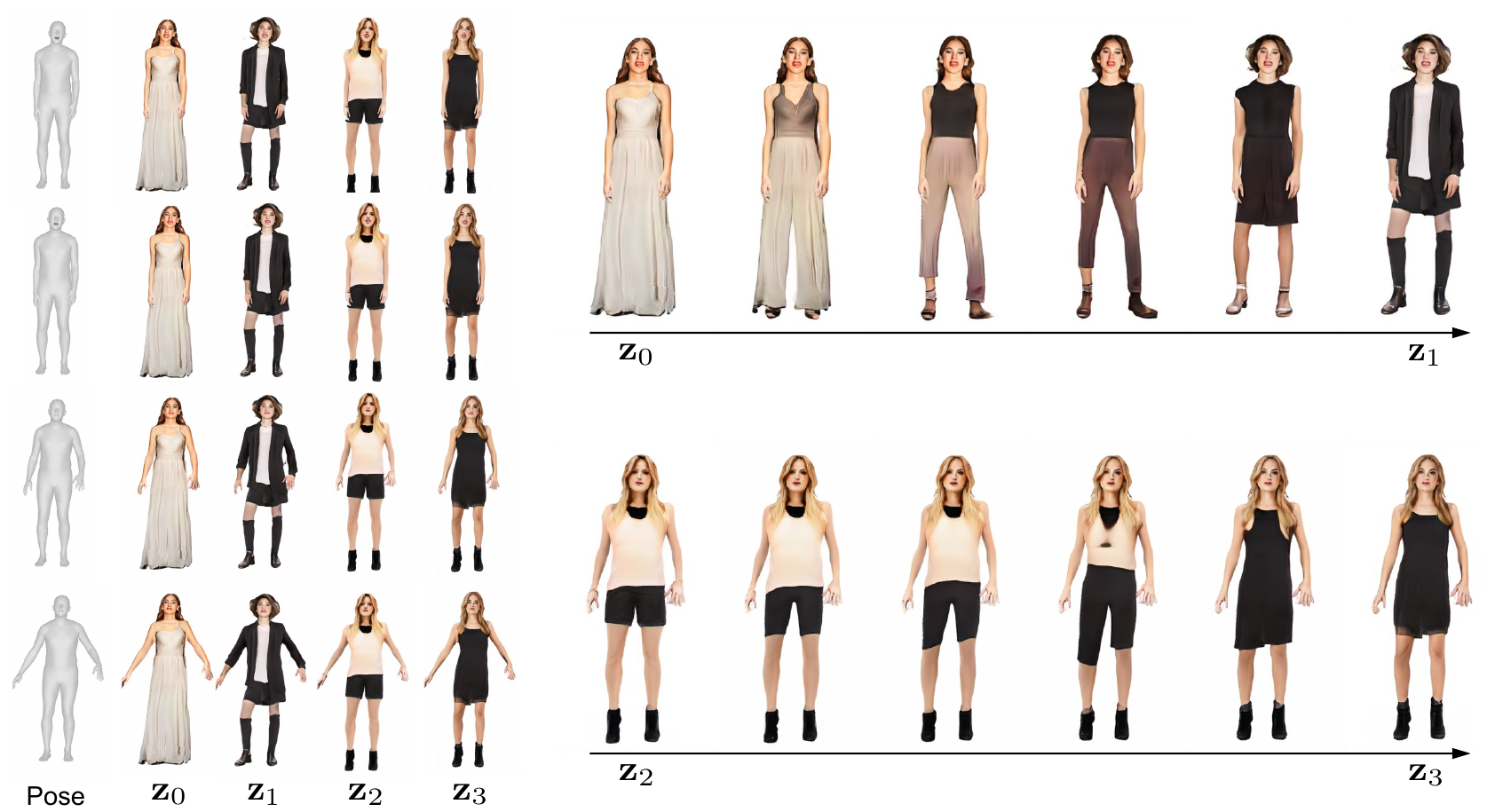}
\caption{Generation results using various latent codes under different jaw and body poses. We do random walk in latent space to demonstrate the disentanglement between identity and condition.}
\label{fig:randomwalk}
\vspace{-3mm}
\end{figure}

{\bf Disentanglement between identity and condition.} Figure~\ref{fig:randomwalk} shows that \ours{} can generate avatars with different identities under the same jaw and pose conditions. We further apply random walk between two latent codes to demonstrate this disentangled generation ability.

\begin{figure}[t]
\centering
\includegraphics[width=\textwidth]{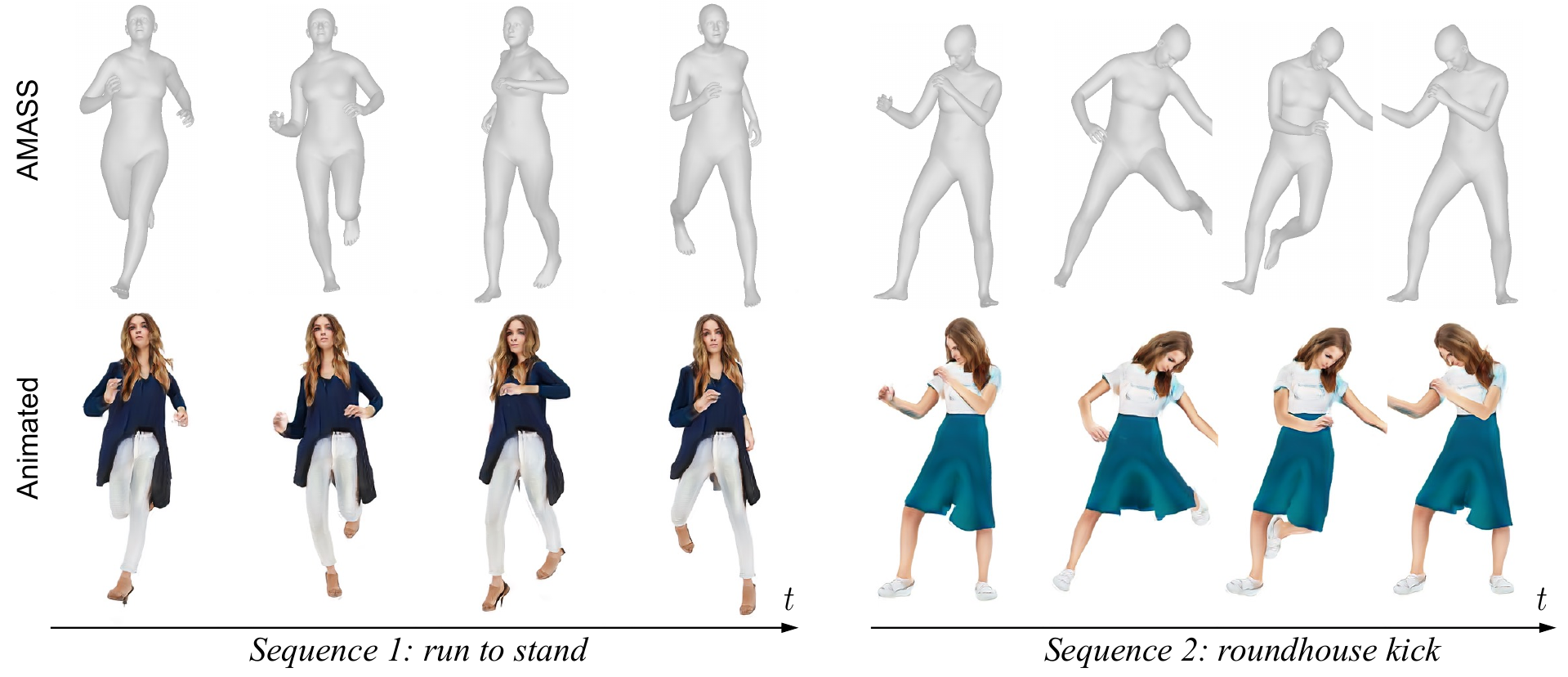}
\caption{Animation results using motion sequences sampled from AMASS~\cite{mahmood2019amass}.}
\label{fig:amass}
\vspace{-3mm}
\end{figure}

{\bf Animation using AMASS.} Figure~\ref{fig:amass} shows more animation results driven by the realistic human motion sequences sampled from AMASS~\cite{mahmood2019amass}. It demonstrates that our generated avatars can be animated by the motion capture results, which could open up possibilities for downstream applications.

\subsection{Additional Ablation Studies}
\begin{table}[t]
  \caption{Additional ablations of our method on SHHQ dataset. We study the effects of SMPL-X errors, discriminator conditions, shared generator, and shared hand Tri-plane.}
  \label{tab:ab}
  \renewcommand{\tabcolsep}{1.8pt}
  \small
\begin{subtable}[!t]{\linewidth}
    \centering
  \begin{tabular}{cccccccccccc}
    \toprule
    {\it SMPL-X}     & FID\(\downarrow\)      &FID\(_\text{f}\)\(\downarrow\)  &FID\(_\text{h}\)\(\downarrow\) &PCK\(\uparrow\)  &PCK\(_\text{f}\)\(\uparrow\) &PCK\(_\text{h}\)\(\uparrow\)&Exp\(\downarrow\) &Shape\(\downarrow\) &Jaw\(\downarrow\) &Body\(\downarrow\) &Hand\(\downarrow\)\\
    \midrule
    noisy &8.61 &10.48 &19.55 &62.17 &90.48 &31.14 &6.78 &3.88&7.45&1.51&3.75\\
    clean &5.88 & 10.06 &19.23 &65.14 &91.44 &38.53 &5.56 &3.66 &6.57 &1.24 &3.30\\
    \bottomrule
  \end{tabular}
  \caption{The effect of SMPL-X estimation errors.}
  \label{tab:ab_smplx}
\end{subtable}
\hspace{\fill}
\begin{subtable}[!t]{0.32\linewidth}
    \centering
  \begin{tabular}{ccccc}
    \toprule
    {\it Expression} &FID\(_\text{f}\)\(\downarrow\) &PCK\(_\text{f}\)\(\uparrow\) &Exp\(\downarrow\)\\
    \midrule
    w/o  &9.57 &91.43 &5.86\\
    w/  &10.06 &91.44 &5.56\\
    \bottomrule
  \end{tabular}
  \caption{The effect of facial expression condition in face discriminator.}
  \label{tab:ab_exp}
\end{subtable}
\hspace{\fill}
\begin{subtable}[!t]{0.32\linewidth}
    \centering
  \begin{tabular}{ccccc}
    \toprule
    {\it Body Pose} &FID\(\downarrow\) &PCK\(\uparrow\) &Body\(\downarrow\)\\
    \midrule
    w/  &14.23 &66.69 &1.00\\
    w/o  &5.88 &65.14 &1.24\\
    \bottomrule
  \end{tabular}
  \caption{The effect of body pose condition in body discriminator.}
  \label{tab:ab_body_pose}
\end{subtable}
\hspace{\fill}
\begin{subtable}[!t]{0.32\linewidth}
    \centering
  \begin{tabular}{ccccc}
    \toprule
    {\it Hand Pose} &FID\(_\text{h}\)\(\downarrow\) &PCK\(_\text{h}\)\(\uparrow\) &Hand\(\downarrow\)\\
    \midrule
    w/o  &19.57 &40.07 &3.59\\
    w/  &19.23 &38.53 &3.30\\
    \bottomrule
  \end{tabular}
  \caption{The effect of hand pose condition in hand discriminator.}
  \label{tab:ab_hand_pose}
\end{subtable}

\begin{subtable}[!t]{0.38\linewidth}
    \centering
  \begin{tabular}{ccccc}
    \toprule
    {\it Generator}     & FID\(\downarrow\)      &FID\(_\text{f}\)\(\downarrow\)  &FID\(_\text{h}\)\(\downarrow\)\\
    \midrule
    separated &7.65 &12.20 &21.03 \\
    shared &5.88 & 10.06 &19.23\\
    \bottomrule
  \end{tabular}
  \caption{The effect of sharing generator branches for body, face, and hand.}
  \label{tab:ab_generator}
\end{subtable}
\hspace{\fill}
\begin{subtable}[!t]{0.6\linewidth}
    \centering
  \begin{tabular}{cccccc}
    \toprule
    {\it Hand Triplanes}     & FID\(\downarrow\)      &FID\(_\text{f}\)\(\downarrow\)  &FID\(_\text{h}\)\(\downarrow\) &PCK\(_\text{h}\)\(\uparrow\)&Hand\(\downarrow\)\\
    \midrule
    double &8.32 &10.12 &20.53 &39.64&3.27\\
    single &5.88 & 10.06 &19.23 &38.53 &3.30\\
    \bottomrule
  \end{tabular}
  \caption{The effect of sharing Tri-planes for two hands.}
  \label{tab:ab_hand_plane}
\end{subtable}

\end{table}

{\bf SMPL-X errors.} \ours{} generates avatars based on the pre-estimated SMPL-X parameters. Therefore, the SMPL-X estimation error could jeopardize the training process. To study its effects, we manually add random noises into the existing SMPL-X estimation results during training. The results reported in Table~\ref{tab:ab_smplx} show that noisy SMPL-X decreases the model's performance in terms of all the evaluation metrics. It mainly affects the control ability and full body image fidelity. Therefore, a more precise SMPL-X estimation method is necessary for improving control ability and full body image quality.

{\bf Discriminator conditions.}
(1) {\it Facial Expression:} The results in Table~\ref{tab:ab_exp} show that although adding expression condition slightly decreases the face image fidelity, it will improve the control ability for facial expression. Thus, we condition the face discriminator on expression parameters.
(2) {\it Body Pose:} It can be observed in Table~\ref{tab:ab_body_pose} that conditioning on body pose will significantly affect image quality. Although it can improve the control ability for body pose, the quality decrease is too large. Thus, we do not condition the discriminator on body pose. We think the reason could be the body pose parameter itself. In SMPL-X, all the poses are encoded by relative angle-axis representations. It is too difficult for the discriminator to decode such abstract geometric transformations. 
(3) {\it Hand Pose:} Table~\ref{tab:ab_hand_pose} summarizes the effect of hand pose condition in discriminator. We can see that although the \(\text{PCK}_\text{h}\) slightly drops by 3.8\%, both the visual quality and the MSE for hand control are improved. Thus, we choose to condition the hand discriminator on hand pose.

{\bf Shared generator.} In Table~\ref{tab:ab_generator}, we study the effects of using separated generator branches for body, face, and hand. It shows that using separated generators cannot improve generation quality. We think the reason would be the redundancy in generators. The redundancy may increase the computation cost and hinder the optimization of the generator.

{\bf Shared hand Tri-plane.} \ours{} utilizes one shared hand Tri-planes for two hands. Table~\ref{tab:ab_hand_plane} shows that using two separated hand Tri-planes can improve the control ability of hand, but image fidelity decreases. We think the reason is that our generator learns to generate an additional hand Tri-plane, which is difficult to optimize. Although independent hand Tri-planes can help model different accessories, accessories usually have small scales. Compared with the full body image, this improvement is less important and will not affect the overall image fidelity too much.

\subsection{Dataset Preprocessing Pipeline}
\label{appendix:process}
The datasets used in our experiments include DeepFashion~\cite{liu2016deepfashion}, MPV~\cite{dong2019towards}, UBC~\cite{zablotskaia2019dwnet}, and SHHQ~\cite{fu2022stylegan}. We follow a specific pipeline to process these datasets, as outlined below:
\begin{enumerate}
    \item \textit{Foreground Mask Estimation}: We employ a segmentation model~\cite{paddleseg2019} to estimate the foreground human body mask for each image. The background is then removed by padding the masked regions with white color.
    \item \textit{Keypoint Detection}: A keypoint estimation model~\cite{alphapose} is used to detect full body keypoints. Based on these keypoints, we crop and align the body, face, and hand images following the method described in~\cite{karras2019style}.
    \item \textit{SMPL-X Parameter Estimation}: We estimate the SMPL-X~\cite{smplx} parameters for the body, face, and hands using a pretrained 3D human reconstruction model~\cite{feng2021collaborative}.
    \item \textit{Image Cropping and Resizing}: The images are cropped once again to move the center joint to the center of the field-of-view. The body image is resized to \(512\times256\), while the face and hand images are resized to \(256^2\).
    \item \textit{Camera Parameters}: We utilize the global orientation and translation from the estimated SMPL-X parameters as the camera extrinsics for each image crop. Additionally, we assume camera focal lengths of \(2560\), \(6400\), and \(8000\) for body, face, and hand camera intrinsics, respectively.
\end{enumerate}
After processing the datasets with the above pipeline, we obtain the following numbers of training images: \(39K\) for SHHQ, \(12K\) for DeepFashion, \(16K\) for MPV, and \(33K\) for UBC. To augment the training data, we follow the approach in~\cite{chan2022efficient} and horizontally flip the images and SMPL-X parameters. 

\subsection{Details for Evaluation Metrics}
\label{appendix:eval}
{\bf Frechet inception distance.} \ours{} generates \(512\times256\) images for full body, which is different from the square images with a resolution of \(512^2\) used in most previous works. We therefore follow EVA3D~\cite{hong2023evad} to pad the images to \(512^2\) to ensure a fair comparison. To compute \(\text{FID}_\text{f}\) and \(\text{FID}_\text{h}\), we leverage the estimated ground truth keypoints derived from the SMPL-X to crop the face and hands regions. This ensures that only the relevant regions are considered in the FID calculation, allowing for a more accurate evaluation of the generated face and hand images.

{\bf Percentage of correct keypoints.} We adopt a pretrained 2D whole body pose estimation model implemented by~\cite{mmpose2020} to estimate 136 keypoints defined by Halpe benchmark~\cite{alphapose}. While previous works~\cite{zhang2023avatargen,hong2023evad,noguchi2022unsupervised} typically use a threshold of 0.5 times the head length to compute PCK, we found that this threshold is too large for the fine-grained keypoints located on the face and hands. Therefore, we use a smaller threshold of 0.1 times the head length to measure the detailed pose control ability. This smaller threshold allows us to assess the accuracy of keypoint localization for these specific regions, providing a more precise evaluation of the pose control capabilities.

{\bf Mean square error for disentangled control abilities.} We follow the evaluation method suggested in prior works~\cite{xu2023omniavatar,deng2020disentangled} to measure the disentangled control abilities. For an expressive human avatar, we evaluate the attributes in \(\{\text{shape}, \text{expression}, \text{jaw pose}, \text{body pose}, \text{hand pose}\}\). When testing each attribute, we randomly sample one novel control parameter from the dataset and replace the parameters for this attribute with the sampled parameter, while the other attributes remain unchanged. Then, we use the 3D human reconstruction model~\cite{feng2021collaborative} to estimated SMPL-X for the synthesized images. In~\cite{xu2023omniavatar,deng2020disentangled}, the control ability is measured by the covariance of the estimated parameters for the selected attribute against the unchanged attributes. However, we find this measurement inaccurate for human body because some works, such as ENARF, cannot produce a clear human body image, leading to inaccurate SMPL-X estimation which causes a large and unstable covariance. Therefore, instead of calculating the covariance, we measure the Mean Square Error (MSE) for each attribute, which can further reflect the fine-grained manipulation ability. In addition, because ENARF and EVA3D are not fully animatable models, we only evaluate body pose for ENARF, and evaluate shape and body pose for EVA3D.

{\bf Mean square error for 3D geometry.} To evaluate the geometry quality quantitatively, we follow the prior work~\cite{zhang2023avatargen} to estimate pseudo ground truth depth map for the generated image. and measure the consistency between pseudo ground truth and the rendered depth map. To measure the depth consistency on both coarse and fine levels, we report Depth for \(128^2\) and Depth256 for \(256^2\). To ensure fair comparisons with previous works, we pad the depth map to a square image format.

{\bf User study.} To compare the generation performance, we conduct a human study for both appearance and geometry. We generate images and the corresponding geometries for four benchmark datasets. We generate 320 samples and recruit 20 participants to measure human preference scores.

\subsection{Details for Baselines}
{\bf ENARF.} We use the official online implementation\footnote{https://github.com/nogu-atsu/ENARF-GAN}. This includes processing the datasets and training the baseline models according to the default settings provided by the authors.

{\bf EVA3D.} For MPV, SHHQ, and UBC datasets, we follow the instructions provided by the authors to process the datasets and estimate SMPL parameters. As for DeepFashion, we use the processed dataset released by the authors. We directly use the checkpoints released by the authors to compute the evaluation metrics for DeepFashion, SHHQ, and UBC. While for MPV, we train EVA3D on it for \(430K\) iterations using the official implementation\footnote{https://github.com/hongfz16/EVA3D}. 

{\bf AvatarGen.} We reproduce AvatarGen model and modified it to condition on SMPL-X parameters, enabling full body controllability. We follow the descriptions in their paper~\cite{zhang2023avatargen} while change their human prior from the observation space to the canonical space to ensure fair comparisons. This also reduces its heavy computation cost for querying SDF values in each iteration.

\subsection{Details for Applications}
{\bf Text-guided avatar synthesis.} For text-guided avatar synthesis, the process involves randomly sampling a latent code \(\mathbf{z}\) and a SMPL-X parameter from the training dataset. The latent code is optimized for \(100\) iterations. While the CLIP~\cite{radford2021learning} loss is used to guide the synthesis process, it may lack supervision for detailed regions such as the face. To address this, a perceptual~\cite{zhang2018perceptual} loss based on the face region is computed to enforce consistency with the image of the original latent code \(\mathbf{z}\). To retarget the synthesized avatar, we randomly sample SMPL-X parameters from an unseen sequence provided by~\cite{yi2022generating}.

{\bf Audio-driven animation.} The generated avatars are driven by the motion sequences sampled from TalkSHOW~\cite{yi2022generating} dataset. However, there is a dimension mismatch between the expression and shape parameters of TalkSHOW and the SMPL-X model we adopt. TalkSHOW uses a dimension of \(100\) for expression and a dimension of \(300\) for shape, while the default SMPL-X model in our approach uses \(50\) parameters for expression and \(200\) parameters for shape. To address this domain gap, we export the body pose, jaw pose, and hand pose from TalkSHOW, while keep the shape and expression parameters unchanged for one avatar. The complete video results for audio-driven animation can be found on our \href{https://showlab.github.io/xagen}{Project Page}.

\section{Broader Impact}
Our model could be used to generate fake individuals or manipulate images for malicious purposes, such as misinformation, harassment, or fraud. These harmful applications could possibly pose a societal threat.

The datasets used to train our model may have inherent biases, such as unbalanced demographic distributions. Therefore, our model may reflect such biases presented in the datasets.  It is essential to be mindful of these biases and consider fairness issues when deploying the model.

To protect our model from misuse and implement safeguards, we will resort to license agreements for model download and usage, which adds restrictions on the access and applications.  

\section{Reproducibility}
In this supplementary material, we provide comprehensive information to ensure the reproducibility of our work. We introduce the implementation details (Section~\ref{appendix:impl}), dataset pre-processing pipeline (Section~\ref{appendix:process}), and details for evaluation metrics (Section~\ref{appendix:eval}). In addition, we provide code for data pre-processing, training, and evaluation.

\bibliography{Reference}